\documentclass[journal]{IEEEtran}
\usepackage{amsmath}
\usepackage{amsfonts}
\usepackage{amssymb}
\usepackage{graphicx}
\usepackage{cite}
\usepackage{float}
\usepackage{multirow}
\usepackage{array}
\usepackage{booktabs}
\usepackage{subfigure}
\usepackage{booktabs}
\usepackage{blindtext}
\usepackage{makecell}
\usepackage{float}
\usepackage{threeparttable}

\usepackage{algpseudocode}
\usepackage{algorithm}

\usepackage[english]{babel}
\usepackage{xcolor}

\newcommand\wu[1]{\textcolor{black}{#1}}

\makeatletter
\renewcommand{\maketag@@@}[1]{\hbox{\m@th\normalsize\normalfont#1}}%
\makeatother

\begin{document}
\newenvironment{sequation}{\begin{equation}\scriptsize}{\end{equation}}
\title{DSTCGCN: Learning Dynamic Spatial-Temporal Cross Dependencies for Traffic Forecasting}

\author{
Binqing Wu, Ling Chen
\thanks{This work was supported by the National Key Research and Development Program of China under Grant 2018YFB0505000. (Corresponding author: Ling Chen.)}
\thanks{Binqing Wu and Ling Chen are with the College of Computer Science and Technology, Zhejiang University, Hangzhou 310027, China (emails: binqingwu@cs.zju.edu.cn, lingchen@cs.zju.edu.cn).}
}

\markboth{$>$ REPLACE THIS LINE WITH YOUR PAPER IDENTIFICATION NUMBER (DOUBLE-CLICK HERE TO EDIT) $<$}%
{Shell \MakeLowercase{\textit{et al.}}: Bare Demo of IEEEtran.cls for IEEE Journals}
\maketitle

\begin{abstract}
Traffic forecasting is essential to intelligent transportation systems, which is challenging due to the complicated spatial and temporal dependencies within a road network. Existing works usually learn spatial and temporal dependencies separately, ignoring the dependencies crossing spatial and temporal dimensions. In this paper, we propose DSTCGCN, a dynamic spatial-temporal cross graph convolution network to learn dynamic spatial and temporal dependencies jointly via graphs for traffic forecasting. Specifically, we introduce a fast Fourier transform (FFT) based attentive selector to choose relevant time steps for each time step based on time-varying traffic data. Given the selected time steps, we introduce a dynamic cross graph construction module, consisting of the spatial graph construction, temporal connection graph construction, and fusion modules, to learn dynamic spatial-temporal cross dependencies without pre-defined priors. Extensive experiments on six real-world datasets demonstrate that DSTCGCN achieves the state-of-the-art performance.
\end{abstract}

\begin{IEEEkeywords}
Traffic forecasting, spatial-temporal graph neural networks, fast Fourier transform
\end{IEEEkeywords}

\IEEEpeerreviewmaketitle

\section{Introduction}
\IEEEPARstart{T}{raffic} forecasting is an essential part of an intelligent transportation system and a crucial technique for developing a smart city \cite{traffic_survey,dl_traffic_survey}. Accurate traffic forecasting will provide reliable guidance for scheduling transportation resources, mitigating traffic congestion, raising early warnings for public safety, and offering suggestions to citizens for their daily commuting \cite{dl_traffic_diff}. Since traffic forecasting has a wide range of real-world applications, it has become a popular research focus in academic and industrial communities for decades.

Traffic forecasting aims to accurately predict future traffic data, e.g., traffic flow and speed, given historical traffic data recorded by sensors on a road network. It is highly challenging due to complicated spatial and temporal dependencies within the road network. Spatially, traffic data collected by a sensor are influenced by nearby traffic conditions, as the traffic dynamics propagate along the road. Temporally, the current features of traffic data are influenced by historical features. Moreover, spatial dependencies and temporal dependencies are entangled and time-varying in real-world traffic systems.

In the past decades, many works have been proposed for this challenging task, from using shallow machine learning  \cite{arima,svr} to applying recurrent neural network (RNN) and convolutional neural network (CNN) based deep learning \cite{dl_traffic_survey,cnn-rnn,cnn-rnn-2,cnn-rnn-3}. Although these works make it possible to model temporal dependencies and grid-based spatial dependencies, they cannot capture graph-based spatial dependencies within an irregular road network in reality \cite{mrabgcn}. Towards this problem, graph neural network (GNN) based works have been proposed to leverage the graph structure of a road network effectively \cite{gnn_survey,traffi_gcn_survey}. Specifically, these works use a graph to define a road network, where each node represents a sensor, and each edge represents a spatial dependency between sensors. More recently, researchers have integrated GNNs to capture spatial dependencies with RNNs \cite{dcrnn,mrabgcn,agcrn}, CNNs \cite{stgcn,astgcn}, or Attentions \cite{gman,astgnn} to model temporal dependencies. This type of network, known as spatial-temporal graph neural networks (STGNNs), has shown the state-of-the-art performance for traffic forecasting \cite{dl_traffic_survey,traffi_gcn_survey}. 

Despite the success, the performance of many existing STGNNs is highly constrained by utilizing static dependencies. They usually construct static graphs, e.g., distance graphs \cite{dcrnn,stgcn,stgode,lv2020temporal}, POI similarity graphs \cite{stmgcn,tfgan}, temporal similarity graphs \cite{stfgcn,tfgan}, and static adaptive graphs \cite{agcrn,mtgnn,rgsl}, to model spatial dependencies, which neglect the changing nature of the spatial dependencies within road networks. Some explorations have been conducted to model such dynamics. For example, a static graph and dynamic attribute based graphs are integrated to obtain time-varying structures \cite{dgcrn}, and attention mechanisms are exploited to construct structures changing with time \cite{dstagnn}. However, these works only focus on the dynamics of spatial dependencies and ignore dependencies crossing spatial and temporal dimensions, which may fail to extract some effective features carried by cross dependencies.

The effectiveness of spatial-temporal cross dependencies has been empirically shown for traffic forecasting. These works \cite{stsgcn,stfgcn,tampsgcnets} usually represent cross dependencies by a fused graph, e.g., a spatial-temporal synchronous graph \cite{stsgcn} constructed by distance graphs and temporal connection graphs, and a spatial-temporal fusion graph \cite{stfgcn} constructed by distance graphs, time similarity graphs, and temporal connection graphs. However, these works still rely on static graphs, which cannot capture dynamic cross dependencies.

To address the aforementioned problems, we propose a \textbf{D}ynamic \textbf{S}patial-\textbf{T}emporal \textbf{C}ross \textbf{G}raph \textbf{C}onvolution \textbf{N}etwork (DSTCGCN). To the best of our knowledge, DSTCGCN is the first work that learns dynamic spatial and temporal dependencies jointly via graphs to explore and utilize time-varying cross dependencies for traffic forecasting. The main contributions of our work are as follows:
\begin{itemize}
    \item Introduce an FFT-based attentive selector to choose the relevant time steps for each time step based on real-world traffic data, which can model the dynamics of temporal dependencies. Moreover, it can limit the temporal neighbors of each time step to a small size and reduce the computational complexity.
    \item Introduce a dynamic cross graph construction module to fuse time-varying spatial graphs and temporal connection graphs in a directed and sparse way, which can model dynamic spatial-temporal cross dependencies without introducing over-fitting problems.
    \item Evaluate DSTCGCN on six real-world datasets for traffic flow and traffic speed forecasting. The comprehensive experimental results demonstrate the state-of-the-art performance of DSTCGCN.
\end{itemize}

\section{Related Work}
\subsection{STGNNs for traffic forecasting}
GNNs have shown superior performance in many applications due to their ability to model non-Euclidean dependencies \cite{gnn_survey}. In particular, for traffic forecasting, STGNNs have shown the state-of-the-art performance, as they can learn spatial dependencies and temporal dependencies more effectively compared with other deep learning works \cite{traffi_gcn_survey}.

Many STGNNs \cite{gnn_survey} integrate GNNs to capture spatial dependencies with RNNs, CNNs, or Attentions to model temporal dependencies. For example, STGCN \cite{stgcn} deploys GCN and 1-D convolution to capture spatial and temporal dependencies, respectively. ASTGCN \cite{astgcn} improves STGCN by introducing spatial and temporal attention mechanisms into the model to capture the dynamics of traffic data. ASTGNN \cite{astgnn} develops a GCN to model the spatial dependencies and a temporal trend-aware multi-head self-attention to capture the temporal dependencies. Since these STGNNs mainly use pre-defined graphs, e.g., geometric distance, functional similarity, and transportation connectivity \cite{stmgcn}, they might miss some implicit spatial dependencies.

To address this problem, some graph learning works have been proposed to construct graph structures from observed data end-to-end. Graph WaveNet \cite{graph_wavenet} learns a self-adaptive adjacency matrix to capture spatial dependencies by multiplying two learnable node embeddings. AGCRN \cite{agcrn} constructs an adjacency matrix directly by multiplying one learnable node embedding and its transpose. MTGNN \cite{mtgnn} learns a uni-directional adjacency matrix using two node embeddings. RGSL \cite{rgsl} further regulates the learned graphs by Gumbel-softmax. Although these learned graphs can alleviate the limitation of pre-defined graphs, they still model static spatial dependencies, which neglect the changing nature of the spatial dependencies within road networks.

There have been some explorations of modeling dynamic spatial dependencies. For example, SLCNN \cite{slcnn} learns dynamic structures by a function of the current samples. DGCRN \cite{dgcrn} constructs dynamic adjacency matrices by integrating dynamic features (time stamps and speeds). DSTAGNN \cite{dstagnn} and D2STGNN \cite{d2stgnn} obtain dynamic adjacency matrices using attention mechanisms. However, these works only focus on the dynamics of spatial dependencies, which ignores time-varying dependencies crossing spatial and temporal dimensions.

\subsection{Cross dependency modeling}
Existing works usually learn spatial and temporal dependencies separately, ignoring the dependencies crossing spatial and temporal dimensions. Recently, some works have started to model such cross dependencies. STSGCN \cite{stsgcn} first proposes a spatial-temporal synchronous graph constructed by spatial graphs and temporal connection graphs to capture the cross dependencies at the adjacent time steps. Following STSGCN, STFGCN \cite{stfgcn} fuses the global temporal similarity graphs with spatial graphs and temporal connection graphs to get a spatial-temporal fusion graph, which extends the time range of cross dependencies to all time steps. TAMP-S2GCNETS \cite{tampsgcnets} constructs directed supra graphs via spatial graphs and temporal connection graphs. AutoSTS \cite{autosts} designs a set of candidate cross graphs for automated spatial-temporal synchronous modeling by neural architecture search algorithm.

However, these works still rely on static graphs, which cannot capture dynamic cross dependencies. More recently, TravseNet \cite{traversenet} unifies all spatial dependencies and temporal dependencies via attention mechanisms. Although TravseNet can model dynamic cross dependencies to some extent, it suffers from too many parameters and has to rely on sparse implementation of deep graph library for computation. Since dynamic cross dependencies are critical for traffic forecasting but not well explored yet, we propose DSTCGCN to capture time-varying cross dependencies by learning dynamic spatial and temporal dependencies jointly without introducing over-fitting problems.

\section{Problem definition}
Following previous studies \cite{dcrnn,mrabgcn}, the task of traffic forecasting is defined as forecasting the future traffic data given the historical traffic data of a road network. Formally, we define these traffic data recorded by sensors located in the road network as a set $\boldsymbol{X}^{1:T}=\{\boldsymbol{X}^1,\boldsymbol{X}^2,…,\boldsymbol{X}^T\} \in \mathbb{R}^{N\times T \times C}$, where $N$ is the total number of time series, $T$ is the input length of each time series, and $C$ is the dimension of input features. $\boldsymbol{X}^t \in \mathbb{R}^{N \times C}$ denotes observed values of $N$ time series at time step $t$. The traffic forecasting task can be formulated as:
\begin{equation}
\widehat{\boldsymbol{X}}^{T+1: T+H}=\mathcal{\boldsymbol{F}}(\boldsymbol{X}^{1:T} ; \boldsymbol{\Theta}) \in \mathbb{R}^{N \times H \times F},
\end{equation}
where $H$ denotes the forecasting horizon, $F$ is the dimension of output features. $\boldsymbol{F}$ is the deep learning network for forecasting. $\boldsymbol{\Theta}$ denotes all learnable parameters of $\mathcal{\boldsymbol{F}}$.

\section{Methodology}
\subsection{Model overview}

\begin{figure*}
    \centering
    \includegraphics[width=0.85 \textwidth]{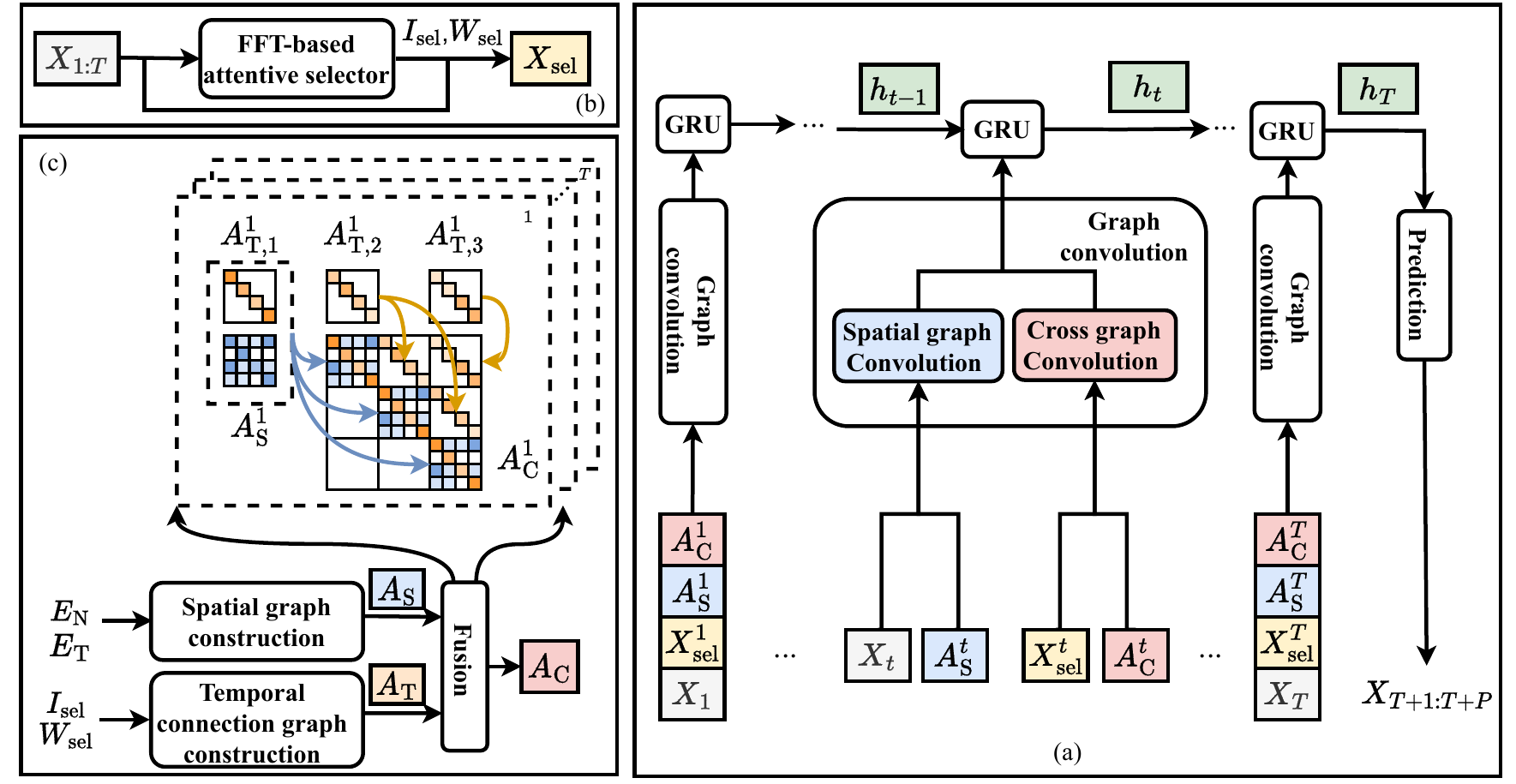}
    \caption{\wu{Architecture of DSTCGCN. (a) Main calculation process of DSTCGCN using GRU as the backbone.} (b) Choosing process of relevant traffic data via the FFT-based attentive selector with the selected indices $\boldsymbol{I}_{\text{sel}}$ and relevant weights $\boldsymbol{W}_{\text{sel}}$. (c) Dynamic cross graph construction process taking the node embedding $\boldsymbol{E}_{\text{N}}$, time embedding $\boldsymbol{E}_{\text{T}}$, $\boldsymbol{I}_{\text{sel}}$, and $\boldsymbol{W}_{\text{sel}}$ as inputs.}
    \label{fig:framework}
\end{figure*}

Learning dynamic spatial and temporal dependencies jointly via graphs is challenging for 1) controlling the parameter number and computational complexity of the model to avoid the risk of over-fitting; and 2) handling the intricate characteristics of different nodes at different time steps. Therefore, we propose DSTCGCN (demonstrated in Fig. \ref{fig:framework}) to tackle these difficulties.

Specifically, an FFT-based attentive selector is introduced to choose relevant time steps for each time step based on time-varying traffic data. Since we renovate the vanilla attention mechanism by FFT and limit the relevant time steps to a small size, the parameter number and computational complexity of the designed selector are well-controlled. A dynamic cross graph construction module is introduced to learn dynamic spatial and temporal dependencies jointly without pre-defined prior, consisting of spatial graph construction, temporal connection graph construction, and fusion parts. Since the constructed temporal connection graphs are diagonal and the cross graphs are upper triangular considering directed information propagation, the graph convolution on the cross graphs does not make the overall model heavy-computing. Moreover, we apply the idea of decomposition to generate parameters of spatial/cross graph convolutions, which can further reduce the model parameters. We utilize GRU as the backbone model and replace the MLP layers in GRU with our graph convolution layers, which can capture spatial, temporal, and spatial-temporal cross dependencies jointly.

\subsection{FFT-based attentive selector}
To alleviate the over-fitting problem and heavy computational burden, we introduce an FFT-based attentive selector. Since attention mechanisms have an excellent ability to deal with dynamic spatial and temporal features \cite{fast_fft,autoformer,zhou2022fedformer}, we design our selector based on an attention mechanism and reduce its computational complexity using FFT.

Drawing insights from instance normalization (IN) \cite{ins_norm} for computer vision tasks, we apply temporal normalization (TN) \cite{st_norm} to extract high-frequency components of traffic data, which can reflect the changing characteristics of inputs. We concatenate them with original traffic data and feed these enriched features into the FFT-based attentive selector. The process can be \wu{formulated as:}
\begin{equation}
\begin{aligned}
    & \boldsymbol{X}_\text{norm}  = \operatorname{TN}(\boldsymbol{X}) \\
    & \tilde{\boldsymbol{X}}  = \operatorname{Concat}(\boldsymbol{X},\boldsymbol{X}_\text{norm}),
\end{aligned}
\end{equation}
where $\boldsymbol{X} \in \mathbb{R}^{N \times T \times C}$, $\tilde{\boldsymbol{X}} \in \mathbb{R}^{N \times T \times 2C}$.

\begin{figure}
    \centering
    \includegraphics[width=0.45\textwidth]{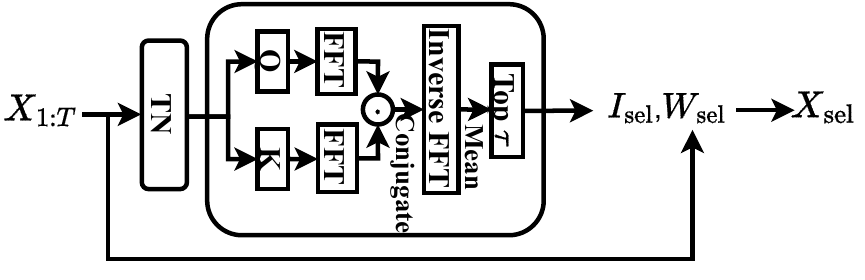}
    \caption{FFT-based attentive selector with temporal normalization.}
    \label{fig:fft}
\end{figure}

As illustrated in Fig. \ref{fig:fft}, the inputs are transformed into query $\boldsymbol{Q}\in \mathbb{R}^{N \times T \times d_{\text{h}}}$ and key $\boldsymbol{K}\in \mathbb{R}^{N \times T \times d_{\text{h}}}$ by linear projections. Inspired by \cite{fast_fft,autoformer,zhou2022fedformer}, applying FFT for attention mechanisms can not only reduce the computational cost but also help to get better temporal representations. Thus, we project $\boldsymbol{Q}$ and $\boldsymbol{K}$ into Fourier space and use Hadamard production instead of matrix multiplication to calculate the attention weights. Formally, the attention weights for the query at time step $i$ $\boldsymbol{Q}^i\in \mathbb{R}^{N \times d_{\text{h}}}$ and the key at time step $j$ $\boldsymbol{K}^j\in \mathbb{R}^{N \times d_{\text{h}}}$ can be \wu{calculated by:}
\begin{equation}
\label{fft}
\begin{aligned}
& \boldsymbol{Q}^i  = \operatorname{Linear}(\tilde{\boldsymbol{X}}^i) \\
& \boldsymbol{K}^j  = \operatorname{Linear}(\tilde{\boldsymbol{X}}^j) \\
& \boldsymbol{M}^{ij} =\mathcal{F}^{-1}((\mathcal{F}(\boldsymbol{Q}^i) \odot \overline{\mathcal{F}(\boldsymbol{K}^j)}),
\end{aligned}
\end{equation}
where $\mathcal{F}$ denotes FFT and $\mathcal{F}^{-1}$ is its inverse. $\overline{(\cdot)}$ is the conjugate operation. $\odot$ is the Hadamard production. $\mathcal{F}(\boldsymbol{Q}^i)\in \mathbb{C}^{N \times d_{\text{F}}}$ and $\mathcal{F}(\boldsymbol{K}^j)\in \mathbb{C}^{N \times d_{\text{F}}}$ are the Fourier transform of $\boldsymbol{Q}^i$ and $\boldsymbol{K}^j$ , respectively, with $d_{\text{F}}<d_{\text{h}}$.

To further reduce the computational complexity and obtain more representative temporal characteristics, the values among the node and feature dimension of $\boldsymbol{M}^{ij}$ are aggregated by mean aggregation to get a single attention value $\boldsymbol{M}_{\text{agg}}^{ij}$ for measuring the relevance between time step $i$ and time step $j$. Given the attention values between all time steps $\boldsymbol{M}_{\text{agg}} \in \mathbb{R}^{T \times T}$, the top-$\tau$ relevant time steps are selected for each time step:
\begin{equation}
\boldsymbol{I}_{\text{sel}},\boldsymbol{W}_{\text{sel}} = \operatorname{Top-\tau} (\boldsymbol{M}_{\text{agg}}),
\end{equation}
where the selected index and relevant weights are denoted as $\boldsymbol{I}_{\text{sel}}\in \mathbb{R}^{T \times \tau}$ and $\boldsymbol{W}_{\text{sel}}\in \mathbb{R}^{T \times \tau}$, respectively. The corresponding relevant traffic data for all time steps are denoted as  $\boldsymbol{X}_{\text{sel}}\in \mathbb{R}^{N \times T \times \tau \times C}$.

\textbf{Computational complexity analysis}: The complexities of FFT, Hadamard production, and mean aggregation are $O(Nd_{\text{F}}log(d_{\text{h}}))$, $O(Nd_{\text{F}})$, and $O(1)$, respectively. Thus, the complexity of calculating the attention value of a pair of time steps (Calculation of $\boldsymbol{M}^{ij}$ in Eq. \ref{fft}) is $O(Nd_{\text{F}}log(d_{\text{h}}))$, while the complexity of the vanilla self-attention to obtain a single attention value is $O((Nd_{\text{h}})^2)$ with the input dimension $Nd_{\text{h}}$.

\subsection{Dynamic cross graph construction}
Learning a graph structure from observed data is promising for pushing forward the forecasting ability of STGNNs, as it breaks the bottleneck of requiring a pre-defined graph structure for GNNs. Recently, some graph learning methods have been proposed. For example, Graph WaveNet \cite{graph_wavenet}, AGCRN \cite{agcrn}, and MTGNN \cite{mtgnn} adopt learnable node embeddings to construct a static graph structure. DAAGCN \cite{daagcn} further utilizes time-varying embeddings to construct dynamic graph structures. Following these previous works, the dynamic graphs are constructed by node and time embeddings.

\subsubsection{Spatial graph construction}
To balance the effectiveness and simplicity of dynamic spatial graph construction, we apply the idea of decomposition. We randomly initialize a node embedding $\boldsymbol{E}_{\text{N}}\in \mathbb{R}^{N \times d_{\text{e}}}$ to represent the node-specific characteristics shared by all time steps, and a time embedding $\boldsymbol{E}_{\text{T}}\in \mathbb{R}^{T \times d_{\text{e}}}$ to indicate the relative dynamic time characteristics for time steps within a time range. We then combine them to learn dynamic spatial graphs. Specifically, for the spatial graph at time step $t$, the construction process can be formulated as:
\begin{equation}
\begin{aligned}
& \boldsymbol{E}^t  =\boldsymbol{E}_{\text{N}} \oplus \boldsymbol{E}_{\text{T}}^t \\
& \boldsymbol{A}_{\text{S}}^t =\operatorname{softmax}(\boldsymbol{E}^t(\boldsymbol{E}^t)^T),
\end{aligned}
\end{equation}
where the time embedding of time step $t$ is denoted as $\boldsymbol{E}^t \in \mathbb{R}^{1 \times d_{\text{e}}}$. $\oplus$ represents the broadcasting addition that adds $\boldsymbol{E}_{\text{T}}^t$ to each row of $\boldsymbol{E}_{\text{N}}$. Therefore, $\boldsymbol{E}^t$ can be regarded as the embedding that integrates the characteristics of nodes and time step $t$. Then, $\boldsymbol{E}^t$ is used to construct the spatial graph at time step $t$ that is denoted as $\boldsymbol{A}_{\text{S}}^t \in \mathbb{R}^{N \times N}$.

\subsubsection{Temporal connection graph construction}
In previous works \cite{stsgcn,stfgcn}, temporal connection graphs are diagonal matrices with shared values for all nodes and all time steps. Such designs make all nodes share the same temporal dependencies for all time steps, which neglects the specific characteristics of nodes and time steps. To address this problem, we construct temporal connection graphs based on the self-dependencies of spatial graphs and the relevant weights of the selected time steps calculated by the FFT-based attentive selector.

For each time step, we construct $\tau$ temporal connection graphs. Taking time step $t$ as an example, the diagonal values of the spatial graph $\boldsymbol{A}_{\text{S}}^t$ are chosen to represent the self-impacts of nodes, denoted as $\boldsymbol{D}_{\text{S}}^t \in \mathbb{R}^{N \times 1}$. Then, the weights of selected time steps $\boldsymbol{W}_{\text{sel}}^t \in \mathbb{R}^{1 \times \tau}$ (calculated by the FFT-based attentive selector) are used as the coefficients to adjust the self-impacts. Since $\boldsymbol{D}_{\text{S}}^t$ and $\boldsymbol{W}_{\text{sel}}^t$ are adaptive to nodes and time steps, respectively, the adjusted self-impacts can be used to construct temporal connection graphs, considering the specific characteristics of nodes and time steps. The learning process can be formulated as:
\begin{equation}
\begin{aligned}
& \boldsymbol{A}_{\text{T},i}^t =\operatorname{fill\_diagonal}(\boldsymbol{D}_{\text{S}}^t \odot \boldsymbol{W}_{\text{sel}, i}^t) \\
& \boldsymbol{A}_{\text{T}}^t =\{\boldsymbol{A}_{\text{T}, t_1}^t, \boldsymbol{A}_{\text{T}, t_2}^t, \ldots, \boldsymbol{A}_{\text{T}, t_\tau}^t\},
\end{aligned}
\end{equation}
where $i \in \{1,2,…,T\}$ indicates the temporal index of the selected time step. $\boldsymbol{A}_{\text{T},i}^t \in \mathbb{R}^{N \times N}$ is the corresponding temporal connection graph. $\boldsymbol{A}_{\text{T}}^t \in \mathbb{R}^{N \times N \times \tau}$ with $t_1 < t_2< \cdots <t_\tau$. It is worth mentioning that $\boldsymbol{D}_{\text{S}}^t$ contains the relative temporal characteristics from the time embedding $\boldsymbol{E}_{\text{T}}^t$ (introduced in (\wu{Section} IV C.1)) and $\boldsymbol{W}_{\text{sel}}^t$ contains the absolute temporal characteristics from $\boldsymbol{X}^t$ (introduced in (\wu{Section} IV A)). Therefore, the constructed temporal connection graphs are time-varying and can model dynamic temporal dependencies, which are more flexible than static temporal connection graphs used in previous works \cite{stsgcn,stfgcn}.

\subsubsection{Fusion}
Given a spatial graph and multiple temporal connection graphs at each time step (both are time-specific), we fuse them into a cross graph. To reduce the computational cost, we follow the assumption that the propagation direction of information is from past to present to future \cite{tampsgcnets}. Thus, the cross graph is directed, and the temporal impact from the time step with a larger index is prioritized during fusing.

\begin{figure}
    \centering
    \includegraphics[width=0.5\textwidth]{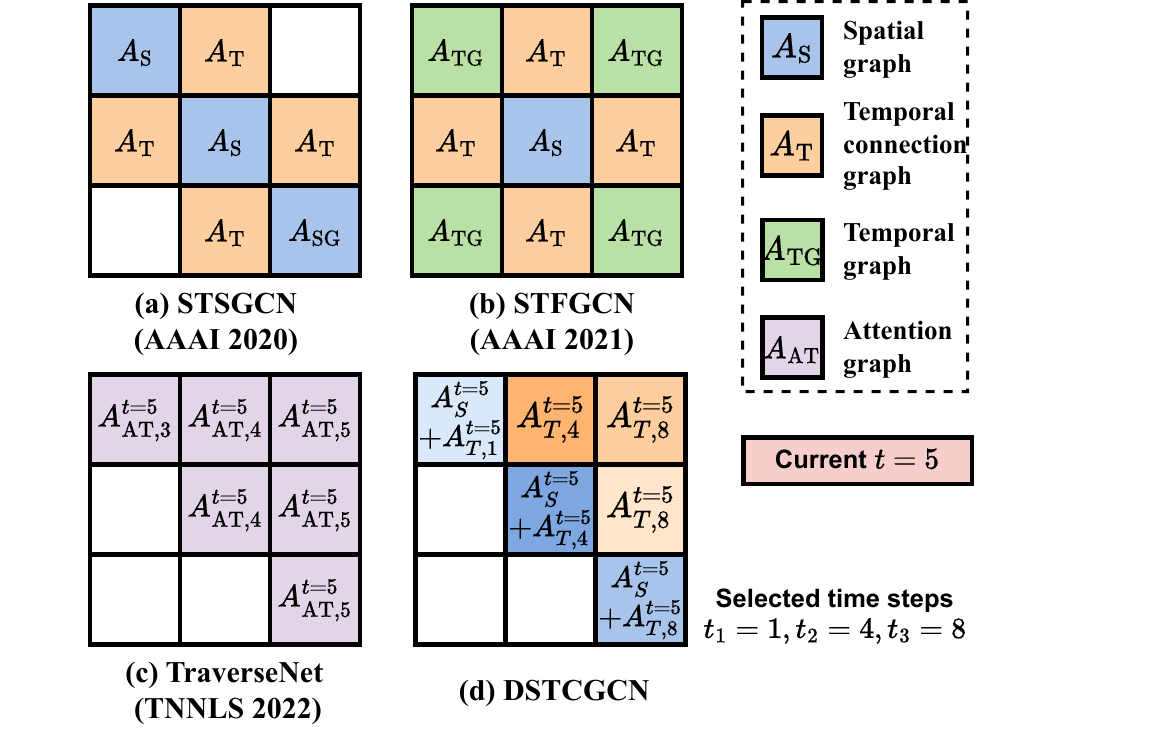}
    \caption{An example to illustrate the cross graphs constructed by different strategies when the time step $t$ is 5 and the number of selected time steps $\tau$ is 3. (a)(b)(c) are the cross graphs from baselines. (d) is the cross graph constructed by our strategy. We assume the selected time steps of our FFT-based selector for $t=5$ are $t_1=1,t_2=4,t_3=8$. The spatial graph $\boldsymbol{A}_{\text{S}}^{t=5}$ is on the diagonal, and the temporal connection graphs 
    $\boldsymbol{A}_{\text{T},1}^{t=5}$, $\boldsymbol{A}_{\text{T},4}^{t=5}$, and $\boldsymbol{A}_{\text{T},8}^{t=5}$ are assigned to corresponding columns of the cross graph.}
    \label{fig:comparison}
\end{figure}

Specifically, for time step $t$, the cross graph $\boldsymbol{A}_{\text{C}}^t \in \mathbb{R}^{\tau N \times \tau N}$ is a directed graph. The spatial graph $\boldsymbol{A}_{\text{S}}^t$ is assigned on the diagonal of the cross graph to maintain the spatial dependencies between nodes. The temporal connection graphs $\boldsymbol{A}_{\text{T}}^t$, regarded as self temporal impacts from selected time steps, are assigned to the upper triangular part of the cross graph. The combination of $\boldsymbol{A}_{\text{S}}^t$ and $\boldsymbol{A}_{\text{T},i}^t$ on the diagonal is the addition for simplicity. Formally, the construction process of the cross graph at time step $t$ can be formalized as:
\begin{equation}
\boldsymbol{A}_{\text{C}}^t=\left[\begin{array}{cccc}
\boldsymbol{A}_{\text{S}}^t+\boldsymbol{A}_{\text{T}, t_1}^t & \boldsymbol{A}_{\text{T}, t_2}^t & \cdots & \boldsymbol{A}_{\text{T}, t_\tau}^t \\
\mathbf{0} & \boldsymbol{A}_{\text{S}}^t+\boldsymbol{A}_{\text{T}, t_2}^t & \cdots & \boldsymbol{A}_{\text{T}, t_\tau}^t \\
\vdots & \vdots & \ddots & \vdots \\
\mathbf{0} & \mathbf{0} & \mathbf{0} & \boldsymbol{A}_{\text{S}}^t+\boldsymbol{A}_{\text{T}, t_\tau}^t
\end{array}\right],
\end{equation}
where $\boldsymbol{A}_{\text{C}}^t \in \mathbb{R}^{\tau N \times \tau N}$ is the cross graph at time step $t$. $t_1 < t_2< \cdots <t_\tau$, and $\tau$ is the number of the selected time steps from the FFT-based selector. An example shown in Fig. \ref{fig:comparison} (d) illustrates the fused cross graph.

\textbf{Comparison with baselines}: Fig. \ref{fig:comparison} gives an example to illustrate the cross graphs constructed by baselines and DSTCGCN. STSGCN \cite{stsgcn} and STFGCN \cite{stfgcn} use static fused graphs for all time steps, ignoring the changing nature of spatial and temporal dependencies within traffic systems. For TraverseNet \cite{traversenet} and \wu{DSTCGCN}, the cross graphs are dynamic. Under the settings of the example, the selected time steps of TraverseNet are three continuous time steps chosen by the fixed rule. In contrast, ours are chosen by the FFT-based attentive selector based on the real-time traffic inputs, which are more flexible. Moreover, since the selecting range is the whole time series, the temporal receptive field of our cross graphs is global, which may handle local outliers such as data missing. In addition, TraverseNet unifies all spatial and temporal dependencies via attention mechanisms, suffering from the high computational cost and the risk of overfitting.

\subsection{Graph convolution}
We introduce the cross graph convolution to extract dynamic spatial and temporal features jointly. Instead of learning the parameters of the cross graph convolution directly, we apply the idea of decomposition to generate them following \cite{agcrn, rgsl}, which can reduce the model parameters. The process can be formalized as:
\begin{equation}
\begin{aligned}
& \boldsymbol{W}_{\text{C}}^t =\boldsymbol{M}^t \boldsymbol{K}_{\text{C}, \text{weights}} \\
& \boldsymbol{B}_{\text{C}}^t =\boldsymbol{M}^t \boldsymbol{K}_{\text{C}, \text{bias}},
\end{aligned}
\end{equation}
where $\boldsymbol{M}^t \in \mathbb{R}^{N \times d_{\text{e}}}$ represents the specific characteristics of each node at each time step. $\boldsymbol{K}_{\text{C, weights}} \in \mathbb{R}^{d_{\text{e}} \times d_{\text{i}} \times d_{\text{o}}}$ ($d_{\text{i}}$ and $d_{\text{o}}$ are the input and out dimensions) and $\boldsymbol{K}_{\text{C, bias}} \in \mathbb{R}^{d_{\text{e}} \times d_{\text{o}}}$ represent the hidden shared pattern for all nodes at all time steps. In particular, as mentioned in \wu{Section} IV C.1, $\boldsymbol{E}^t$ can be regarded as the matrix that integrates the characteristics of nodes and time step $t$. Thus, we utilize $\boldsymbol{E}^t \in \mathbb{R}^{N \times d_{\text{e}}}$ as the character matrices $\boldsymbol{M}^t$ when generating parameters for time step $t$ directly.

The cross graph convolution based on the message passing theory can be formalized as:
\begin{equation}
\boldsymbol{H}_{\text{C}, \text{out}}^t=\sigma((\boldsymbol{A}_{\text{C}}^t+\boldsymbol{I}) \boldsymbol{H}_{\text{C}, \text{in}}^t \boldsymbol{W}_{\text{C}}^t+\boldsymbol{B}_{\text{C}}^t),
\end{equation}
where $\boldsymbol{H}_{\text{C,in}}^t \in \mathbb{R}^{\tau \times N \times d_{\text{i}}}$ and $\boldsymbol{H}_{\text{C,out}}^t \in \mathbb{R}^{\tau \times N \times d_{\text{o}}}$ are the input and output of cross graph convolution. For the first layer, $\boldsymbol{H}_{\text{C,in}}^t$ is $\boldsymbol{X}_{\text{sel}}^t \in \mathbb{R}^{\tau \times N \times C}$. $\boldsymbol{I} \in \mathbb{R}^{\tau N \times \tau N}$ is the identity matrix, which represents the self-loop. We also introduce the spatial convolution following the same design of the cross graph convolution to preserve the pure spatial dependencies at the time step $t$, and extract the current features denoted as $\boldsymbol{H}_{\text{S,out}}^t \in \mathbb{R}^{N \times d_{\text{o}}}$. Then, $\boldsymbol{H}_{\text{C,out}}^t$ and $\boldsymbol{H}_{\text{S,out}}^t$ are fused as the final output of graph convolution module, which can be formulated as:
\begin{equation}
\boldsymbol{H}_{\text{G}, \text {out}}^t=\operatorname{Linear}(\operatorname{Pooling}(\boldsymbol{H}_{\text{C}, \text {out}}^t), \boldsymbol{H}_{\text{S}, \text {out}}^t),
\end{equation}
where $\boldsymbol{H}_{\text{G},\text {out}}^t \in \mathbb{R}^{N \times d_{\text{o}}}$. $\operatorname{Pooling}(\boldsymbol{H}_{\text{C}, \text{out}}^t)$ denotes the pooling operation on $\boldsymbol{H}_{\text{C}, \text{out}}^t$ to obtain the representive features for $\tau$ time steps. The shape of the corresponding output is the same as $\boldsymbol{H}_{\text{S}, \text{out}}$.

\subsection{Forecasting module}
We utilize GRU as the backbone and replace the MLP layers in GRU with our graph convolution layers as \cite{agcrn, rgsl}. The forward propagation equations for time step $t$ are as follows:
\begin{equation}
\begin{aligned}
    & \boldsymbol{z}^t =\sigma(\operatorname{Gonv}_{\text{z}}(\boldsymbol{X}^t, \boldsymbol{X}_{\text{sel}}^t, \boldsymbol{h}^{t-1})) \\
    & \boldsymbol{r}^t =\sigma(\operatorname{Gonv}_{\text{r}}(\boldsymbol{X}^t, \boldsymbol{X}_{\text {sel}}^t, \boldsymbol{h}^{t-1})) \\
    & \boldsymbol{c}^t =\tanh(\operatorname{Gonv}_{\text{c}}(\boldsymbol{X}^t, \boldsymbol{X}_{\text{sel}}^t,(\boldsymbol{r}^t \odot \boldsymbol{h}^{t-1}))) \\
    & \boldsymbol{h}^t =\boldsymbol{z}^t \odot \boldsymbol{h}^{t-1}+(1-\boldsymbol{z}^t) \odot \boldsymbol{c}^t,
\end{aligned}
\end{equation}
where $\boldsymbol{X}^t \in \mathbb{R}^{N \times C}$ denotes the traffic data at time step $t$. $\boldsymbol{X}_{\text{sel}}^t$ is the traffic data of the selected time steps. We take the final hidden state $\boldsymbol{h}^T$ to predict traffic data of all nodes for the next $H$ steps by a linear transformation, which decreases the time consumption and avoids cumulative error caused by sequential forecasting.

The loss function in this work is L1 loss function:
\begin{equation}
\mathcal{L}(\boldsymbol{\Theta})=\sum_{t=T+1}^{T+H}|\boldsymbol{X}^t-\widehat{\boldsymbol{X}^t}|,
\end{equation}
where $\Theta$ denotes all the learnable parameters in DSTCGCN. $\boldsymbol{X}^t$ and $\widehat{\boldsymbol{X}^t}$ are the ground truth and the forecasting results, respectively.

\section{Experiments}
\subsection{Dataset}
We evaluate DSTCGCN on six public traffic datasets, including PEMS03/4/7/8 provided by \cite{stsgcn,astgnn} and METR-LA/PESM-BAY provided by \cite{dcrnn}, and follow the corresponding suggested data preprocessing strategies. Specifically, we fill the missing data by linear interpolation and aggregate data into 5-minute intervals, resulting in 12 time steps per hour. Besides, we normalize all datasets using the Z-Score normalization. The detailed statistics are shown in Table \ref{table:dataset_statistics}. In the “Signals” column, F represents traffic flow, S represents traffic speed, and O represents traffic occupancy rate. For PEMS03/4/7/8, we only use the traffic flow in the following experiments.

\begin{table}[]
    \caption{Dataset statistics.}
    \resizebox{0.5\textwidth}{!}{
    \centering
    \begin{tabular}{c|c|c|c|c}
        \hline
        Datasets & \#Nodes & TimeRange & \#TimeSteps & Signals \\
        \hline
        PEMS03 & 358 & 9/1/2018 - 11/30/2018 & 26,208 & F \\
        PEMS04 & 307 & 1/1/2018 - 2/28/2018 & 16,992 & F,S,O \\
        PEMS07 & 883 & 5/1/2017 - 8/31/2017 & 28,224 & F \\
        PEMS08 & 170 & 7/1/2016 - 8/31/2016 & 17,856 & F,S,O \\
        METR-LA & 207 & 3/1/2012 - 6/30/2012 & 34,272 & S \\        
        PEMS-BAY & 325 & 1/1/2017 - 5/31/2017 & 52,116 & S \\        
        \hline
    \end{tabular}}
    \label{table:dataset_statistics}
\end{table}

\subsection{Experimental settings}
We split datasets chronologically for training, validation, and testing with the ratio 6:2:2 for PEMS03/4/7/8 and 7:1:2 for METR-LA/PESM-BAY. For our traffic forecasting task, one-hour data are used as input to forecast the next hour’s data as output. Each hour has 12 continuous time steps. DSTCGCN is implemented in Python with PyTorch 1.9.0 and trained on an NVIDIA GeForce RTX 3080 Ti GPU card using Adam optimizer with an initial learning rate of 0.003 and batch size of 64. Neural Network Intelligence (NNI) toolkit is applied to tune important hyperparameters automatically, which can reduce computational costs efficiently. The search space of important hyperparameters and their final choices for different datasets are summarized in Table \ref{table:hyperparameter_settings}. The codes are available at https://github.com/water-wbq/DSTCGCN/.

\begin{table*}[]
\caption{The search spaces of important hyperparameters and their final choices.}
\centering
\begin{tabular}{ll|cccccc}
\hline
\multirow{2}{*}{Hyperparameters} & \multirow{2}{*}{Considered values} & \multicolumn{6}{c}{Dataset}\\ \cline{3-8} 
& & PEMS03 & PEMS04 & PEMS07 & PEMS08 & METR-LA & PEMS-BAY \\
\hline
Node embedding dimension                & \{2, 4, 6, 8, 10, 12\}   & 12 & 10 & 6 & 12 & 10 & 8  \\
\makecell[l] {Hidden dimension of FFT-based attentive selector} & \{8, 16, 32, 64\}  & 32 & 16 & 32 & 8 & 32 & 16\\
Number of graph convolution layers      & \{1, 2, 3\}               & 1 & 2 & 1 & 2 & 2 & 2   \\
Number of RNN units of GRU              & \{16, 32, 64, 128\}      & 64 & 64 & 32 & 64 & 64 & 32 \\
Number of relevant time steps           & \{1, 2, 3, 4\}          & 3 & 3 & 3 & 3 & 2 & 3  \\ 
\hline
\end{tabular}
\label{table:hyperparameter_settings}
\end{table*}

The evaluation metrics, including Mean Absolute Error (MAE), Root Mean Squared Error (RMSE), and Mean Absolute Percentage Error (MAPE) on test data, are reported based on the saved best model that runs for 100 epochs on the validation data.
\begin{equation}
\begin{aligned}
    & \text{MAE} =\frac{1}{H} \sum_{t=T+1}^{T+H}|\boldsymbol{X}^t-\widehat{\boldsymbol{X}^t}| \\
    & \text{RMSE} =\sqrt{\frac{1}{H} \sum_{t=T+1}^{T+H}(\boldsymbol{X}^t-\widehat{\boldsymbol{X}^t})^2} \\
    & \text{MAPE} =\frac{1}{H} \sum_{t=T+1}^{T+H}|\frac{\boldsymbol{X}^t-\widehat{\boldsymbol{X}^t}}{\boldsymbol{X}^t}|.
\end{aligned}
\end{equation}

\subsection{Baselines}

\begin{table*}[]
\caption{Forecasting performance comparison of models on PEMS03/4/7/8.}
\label{all_res}
\centering
\resizebox{\textwidth}{!}{
\begin{tabular}{c|ccc|ccc|ccc|ccc}
\hline
\multirow{2}{*}{Model} & \multicolumn{3}{c|}{PEMS03} & \multicolumn{3}{c|}{PEMS04} & \multicolumn{3}{c|}{PEMS07} & \multicolumn{3}{c}{PEMS08} \\ \cline{2-13} 
                      & MAE     & RMSE    & MAPE   & MAE     & RMSE    & MAPE   & MAE     & RMSE    & MAPE   & MAE     & RMSE    & MAPE   \\ \hline
HA                     & 31.58   & 52.39   & 33.78  & 38.03   & 59.24   & 27.88  & 45.12   & 65.64   & 24.51  & 34.86   & 59.24   & 27.88  \\
ARIMA (J Forecast, 1997)                 & 35.41   & 47.59   & 33.78  & 33.73   & 48.80    & 24.18  & 38.17   & 59.27   & 19.46  & 31.09   & 44.32   & 22.73  \\
VAR (S-PLUS, 2006)& 23.65 & 38.26 & 24.51 & 24.52& 38.61& 17.24 & 50.22& 75.63& 32.22 & 19.19& 29.81& 13.10 \\
FC-LSTM (NeurIPS 2015)  & 21.33   & 35.11   & 23.33  & 26.77   & 40.65   & 18.23  & 29.98   & 45.94   & 13.20   & 23.09   & 35.17   & 14.99  \\
STGCN (IJCAI 2018)                  & 17.55   & 30.42   & 17.34  & 21.16   & 34.89   & 13.83  & 25.33   & 39.34   & 11.21  & 17.50    & 27.09   & 11.29  \\
ASTGCN (AAAI 2019)             & 17.34   & 29.56   & 17.21  & 22.93   & 35.22   & 16.56  & 24.01   & 37.87   & 10.73  & 18.25   & 28.06   & 11.64  \\
GraphWaveNet   (IJCAI 2019)        & 19.12   & 32.77   & 18.89  & 24.89   & 39.66   & 17.29  & 26.39   & 41.50    & 11.97  & 18.28   & 30.05   & 12.15  \\
AGCRN   (NeurIPS 2020)               & 15.98   & 28.25   & 15.23  & 19.83   & 32.26   & 12.97  & 22.37   & 36.55   & 9.12   & 15.95   & 25.22   & 10.09  \\
MTGNN     (KDD 2020)             & \underline{15.10}    & 25.93   & 15.67  & 19.32   & 31.57   & 13.52  & 22.07   & 35.80    & 9.21   & 15.71   & \underline{24.62}   & 10.03  \\
StemGNN   (NeurIPS 2020)              & 18.11   & 30.53   & 16.95  & 21.61   & 33.80    & 16.10   & 22.23   & 36.46   & 9.20    & 15.91   & 25.44   & 10.90   \\
Z-GCNETs  (ICML 2021)             & 16.64   & 28.15   & 16.39  & 19.50    & 31.61   & 12.78  & 21.77   & 35.17   & 9.25   & 15.76   & 25.11   & 10.01  \\
RGSL (IJCAI 2022) & 15.65 & 27.98 & \underline{14.67}  & \underline{19.19} & \underline{31.14} & \underline{12.69} & 20.73 & 34.48 & \underline{8.71} & \underline{15.49} & 24.80 & 9.96 \\
DGCRN (TKDD 2022) & 15.98& 27.41& 17.73 & 20.39& 32.34& 14.64& \underline{20.52} & \underline{33.56} & 9.09 &16.22 &26.10  & 12.06\\
DSTAGNN   (ICML 2022)             & 15.57   & 27.21   & 14.68  & 19.30    & 31.46   & 12.70   & 21.42   & 34.51   & 9.01   & 15.67   & 24.77   & \underline{9.94}   \\
STSGCN (AAAI 2020)                 & 17.48   & 29.21   & 16.78  & 21.19   & 33.65   & 13.90   & 24.26   & 39.03   & 10.21  & 17.13   & 26.80    & 10.96  \\
STFGNN (AAAI 2021)                 & 16.77   & 28.34   & 16.30   & 20.48   & 32.51   & 16.77  & 23.46   & 36.60    & 9.21   & 16.94   & 26.25   & 10.60   \\
TraverseNet (TNNLS 2022)            & 15.44   & \textbf{24.75}   & 16.41  & 19.86   & 31.54   & 14.38  & 21.99       & 35.83       & 9.45      & 15.68   & \underline{24.62}   & 10.87  \\ \hline
DSTCGCN (Ours)                  & \textbf{14.90}    & \underline{25.79}    & \textbf{14.07}  & \textbf{19.01}   & \textbf{31.03}   & \textbf{12.64} & \textbf{20.37}   & \textbf{33.47}   & \textbf{8.64}   & \textbf{15.18}   & \textbf{24.49}   & \textbf{9.68}   \\ \hline
\end{tabular}
}
\end{table*}

\begin{table*}[]
\caption{Forecasting performance comparison of models based on graph learning on METR-LA/PEMS-BAY.}
\label{metr-bay-adap}
\centering
\resizebox{\textwidth}{!}{%
\begin{tabular}{c|ccc|ccc|ccc|ccc|ccc|ccc}
\hline
\multirow{3}{*}{Model} & \multicolumn{9}{c|}{METR-LA} & \multicolumn{9}{c}{PEMS-BAY} \\ 
\cline{2-19}
& \multicolumn{3}{c|}{Horizon 3} & \multicolumn{3}{c|}{Horizon 6} & \multicolumn{3}{c|}{Horizon 12}  & \multicolumn{3}{c|}{Horizon 3} & \multicolumn{3}{c|}{Horizon 6} & \multicolumn{3}{c}{Horizon 12}\\
\cline{2-19}
& MAE & RMSE & MAPE  & MAE & RMSE & MAPE & MAE & RMSE & MAPE  & MAE & RMSE & MAPE  & MAE & RMSE & MAPE & MAE & RMSE & MAPE \\
\hline
AGCRN (S) &2.87 & 5.58 & 7.70 & 3.23 & 6.58 & 9.00 & 3.62 & 7.51 & 10.38 & 1.37 & 2.87 & 2.94 & 1.69 & 3.85 & 3.87 & 1.96 & 4.54 & 4.64 \\
MTGNN (S) &2.69 & 5.18 & 6.86 & 3.05 & 6.17 & 8.19 & 3.49 & 7.23 & 9.87 & 1.32 & 2.79 & 2.77 & 1.65 & 3.74 & 3.69 & 1.94 & 4.49 & 4.53 \\
RGSL (S) & 2.66 & 5.10 & 6.66 & 3.02 & 6.10 & 8.07 & \textbf{3.36} & \underline{7.16} & \textbf{9.68} & \underline{1.29} & 2.73 & 2.72 & \underline{1.62} & 3.64 & 3.57 & \underline{1.85} & \underline{4.32} & \underline{4.39} \\
DGCRN (D) & \underline{2.62} & \underline{5.01} & \underline{6.63} & \underline{2.99} & \underline{6.05} & \underline{8.02} & 3.44 & 7.19 & 9.73 & 1.28 & \underline{2.69} & \underline{2.66} & \textbf{1.59} & \underline{3.63} & \underline{3.55} & 1.89 & 4.42 & 4.43 \\
\hline
DSTCGCN (D) & \textbf{2.57} & \textbf{4.92} & \textbf{6.39} & \textbf{2.96} & \textbf{5.94} & \textbf{7.94} & \underline{3.42} & \textbf{7.15} & \underline{9.70} & \textbf{1.26} & \textbf{2.67} & \textbf{2.64} & 1.63 & \textbf{3.60} & \textbf{3.51} & \textbf{1.84} & \textbf{4.30} & \textbf{4.34} \\
\hline
\end{tabular}}
\end{table*}

The baselines used for our comparative evaluation can be divided into four categories. The first category is classic time-series forecasting methods, including HA, ARIMA \cite{arima} with Kalman filter, VAR \cite{var}, and FC-LSTM \cite{fc_lstm}.

The second category is STGNNs with static graphs with two sub-categories:
\begin{itemize}
    \item With static pre-defined adjacency matrice:
    \begin{itemize}
        \item STGCN \cite{stgcn} combines graph convolutional and two temporal gated convolutions to capture spatial and temporal dependencies.
        \item ASTGCN \cite{astgcn} applies spatial attention to model spatial dependencies between different locations and temporal attention to capture the dynamic temporal dependencies between different time steps.
    \end{itemize}
    \item With static adaptive adjacency matrice:
    \begin{itemize}
        \item Graph WaveNet \cite{graph_wavenet} constructs a self-adaptive adjacency matrix using two node embeddings dictionaries and applies graph convolution with dilated casual convolution to capture spatial and temporal dependencies.
        \item AGCRN \cite{agcrn} generates an adaptive adjacency matrix using one node embedding and uses node adaptive graph convolution and GRU to capture node-specific spatial and temporal dependencies, respectively.
        \item MTGNN \cite{mtgnn} learns a uni-directional adjacency matrix using two node embeddings and uses GNN and dilated convolution for multi-variate time series forecasting.
        \item StemGNN \cite{stemgnn} embeds the inputs using GRU and treats the attention score of the last hidden state as the adjacency matrice. It then combines graph Fourier transform to model inter-series correlations and discrete Fourier transform to model temporal dependencies.
        \item Z-GCNETs \cite{z_gcnnets} integrates the new time-aware zigzag topological layer into time-conditioned GCNs for modeling spatial and temporal dependencies.
        \item RGSL \cite{rgsl} regularizes the adaptive adjacency matrix learned by node embeddings using Gumble-softmax. It applies GNN and GRU to model spatial and temporal dependencies, respectively.
    \end{itemize}
\end{itemize}

The third category is STGNNs with dynamic graphs, including:
\begin{itemize}
    \item DCGRN \cite{dgcrn} constructs the adjacency matrix using static node embeddings and integrates dynamic features (time stamps and speeds) to adjust the matrix for modeling dynamic spatial dependencies.
    \item DSTAGNN \cite{dstagnn} obtains the adjacency matrix by calculating the cosine similarity of input traffic data and uses attention mechanisms to adjust the matrix for dynamics.
\end{itemize}

The fourth category is STGNNs considering spatial-temporal cross dependencies, including:
\begin{itemize}
    \item STSGCN \cite{stsgcn} utilizes localized spatial-temporal subgraphs to capture the localized spatial-temporal dependencies synchronously.
    \item STFGCN \cite{stfgcn} assembles a gated dilated CNN module with a spatial-temporal fusion graph module in parallel to capture local and global dependencies simultaneously.
    \item TraverseNet \cite{traversenet} utilizes attention to capture spatial and temporal dependencies.
\end{itemize}

\subsection{Overall Comparison}
Table \ref{all_res} shows the average forecasting performance over 12 horizons of DSTCGCN and other baselines on four traffic flow datasets, i.e, PEMS03/4/7/8. \textbf{Bolt} denotes the best performance of each metric and \underline{Underline} is the second best. The following phenomena can be observed:

\begin{itemize}
    \item Static graph-based methods (second category) perform better than those of the classic time-series forecasting models (first category) significantly, as they can capture static spatial dependencies between traffic data. Moreover, static adaptive graph-based methods (Graph WaveNet, AGCRN, MTGNN, StemGNN, Z-GCNETs, and RGSL) outperform pre-definded graph-based methods (ST-GCN and ASTGCN). One of the possible reasons for this phenomenon is that learning graphs from historical observed traffic data can discover some underlying spatial dependencies.
    \item Dynamic graph-based methods (third category) and cross graph-based methods (fourth category) have an average superior performance than static graph-based methods due to their abilities of modeling dynamics and cross dependencies, respectively.
    \item DSTCGCN (ours) outperforms almost all baselines on MAE, RMSE, and MAPE over four datasets, achieving a new state-of-the-art traffic flow forecasting performance. Specifically, DSTCGCN yields an average 11.47\% relative MAPE reduction on four traffic flow datasets, benefiting from modeling dynamic cross dependencies.
\end{itemize}

Since the performance of some graph learning methods, including static adaptive graph-based methods (AGCRN, MTGNN, and RGSL) and dynamic adaptive graph-based method (DGCRN), are also promising in Table \ref{all_res}, we further compare DSTCGCN with them on two traffic speed datasets, i.e., METR-LA and PEMS-BAY, on 3, 6, 9, and 12 horizons. The results are in Table \ref{metr-bay-adap}, where S and D mean static and dynamic graph-based methods, respectively. Table \ref{metr-bay-adap} shows that DSTCGCN achieves 14 best cases out of 18 cases. One of the possible reasons is that these four baselines all focus on learning underlying spatial graphs (static or dynamic), neglecting to model temporal connection graphs. Therefore, DSTCGCN, which considers dynamic spatial and temporal connection graphs jointly, performs better than these methods.

\subsection{Ablation study}

\begin{table}[]
\caption{Effects of different components on PEMS04 (average 12 horizions) and METR-LA (12th horizion).}
\label{ablation_study}
\centering
\begin{tabular}{c|ccc|ccc}
\hline
\multirow{2}{*}{Variant} & \multicolumn{3}{c|}{PEMS04}                       & \multicolumn{3}{c}{METR-LA}                   \\
\cline{2-7}
                         & MAE            & RMSE           & MAPE           & MAE           & RMSE          & MAPE          \\
\hline
w/o FFT-AS-1             & 19.62          & 31.67          & 12.87          & 3.62          & 7.33          & 10.12         \\
w/o FFT-AS-2             & 19.26          & 31.15          & 12.62          & 3.48          & 7.22          & 10.05         \\
w/o   TN                 & 19.15          & 31.84          & 12.71          & 3.45          & 7.41          & 10.10          \\
w/o DSG                  & 19.53          & 31.92          & 13.15          & 3.61          & 7.28          & 10.28         \\
w/o   DTCG                & 19.36          & 31.83          & 12.46          & 3.59          & 7.42          & 10.53          \\
w/o   DSTCG               & 19.84          & 32.27          & 12.85          & 3.77          & 7.40          & 10.36         \\
\hline
DSTCGCN                 & \textbf{19.01} & \textbf{31.03} & \textbf{12.64} & \textbf{3.42} & \textbf{7.15} & \textbf{9.70} \\
\hline
\end{tabular}
\end{table}

To verify the effectiveness of each component of DSTCGCN, we conduct the ablation studies with the following variants on PEMS04 and METR-LA:
\begin{itemize}
    \item Without the FFT-based attentive selector (w/o FFT-AS): (1) select $\tau$ time steps as relevant time steps randomly; (2) select $\tau$ latest time steps as relevant time steps.
    \item Without temporal normalization (w/o TN): select the relative time steps only based on the pure raw traffic data without temporal normalized data.
    \item Without dynamic spatial graphs (w/o DSG): replace the dynamic spatial graphs with the static spatial graphs calculated by $\boldsymbol{A}_{\text{S}}=\operatorname{softmax}(\boldsymbol{E}_{\text{N}} (\boldsymbol{E}_{\text{N}})^T)$.
    \item Without dynamic temporal connection graphs (w/o DTCG): replace the temporal connection graphs with identity diagonal matrices on the upper triangular parts of spatial-temporal cross graphs.
    \item Without dynamic spatial-temporal cross graphs (w/o DSTCG): only use dynamic spatial graphs to model the spatial dependencies. 
\end{itemize}

The corresponding results are shown in Table \ref{ablation_study}. The performance of DSTCGCN is better than \wu{those of other variants}, which confirms the effectiveness of each component in our model. We also observe that the forecasting accuracy of the variant w/o DSTCG drops most compared to others, which verifies the significance of modeling dynamic spatial and temporal dependencies jointly via graphs for real-world traffic flow and speed forecasting. In addition, the performance of the variants w/o DSG and w/o DTCG degrades obviously, showing the effectiveness of modeling dynamics

Moreover, since the variant w/o FFT-AS-1 selects $\tau$ time steps randomly, it ignores the temporal dependencies between time steps and leads to a performance drop. On the other hand, the variant w/o FFT-AS-2, considering the latest $\tau$ time steps, has a slight decrease in performance, which may suffer from some local noise. These results verify the effectiveness of our designed FFT-based attentive selector experimentally, as it can select the relevant time steps based on the similarity of real-world traffic data.

\subsection{Hyperparameter Evaluation}

\begin{figure}[]
\centering
\subfigure[Evaluation of $d_e$]{
\begin{minipage}[t]{0.45\linewidth}
\centering
\includegraphics[width=0.95\linewidth]{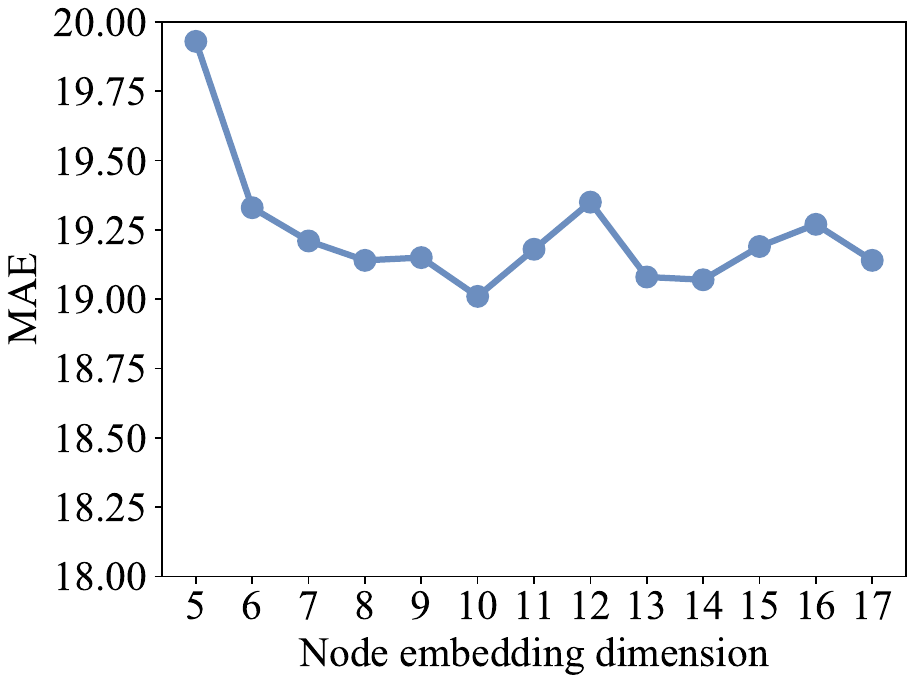}
\label{dim_04}
\end{minipage}
}
\subfigure[Evaluation of $\tau$]{
\begin{minipage}[t]{0.45\linewidth}
\centering
\includegraphics[width=0.95\linewidth]{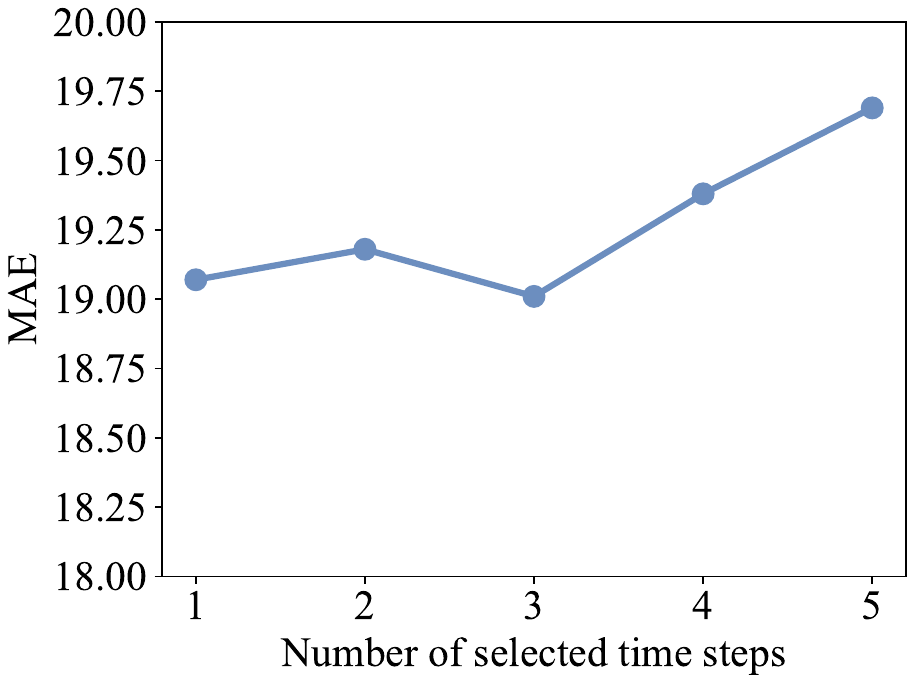}
\end{minipage}
\label{tau_04}
}
\caption{Effects of the dimension of the node and time embeddings $d_e$ and the number of the selected time steps $\tau$ on PEMS04.}
\end{figure}

\begin{figure}[]
\centering
\subfigure[MAE on PEMS04]{
\begin{minipage}[t]{0.45\linewidth}
\centering
\includegraphics[width=1\linewidth]{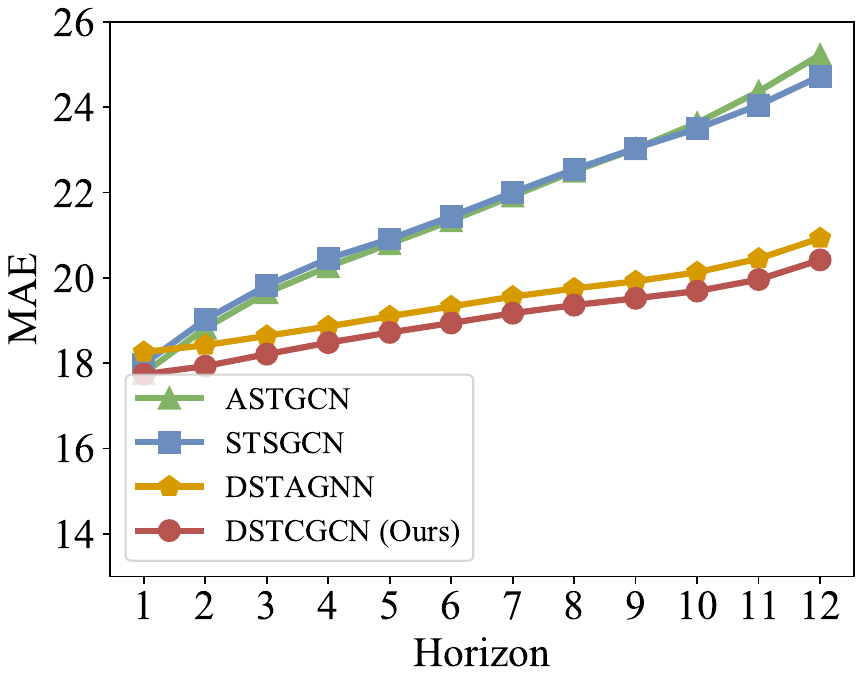}
\end{minipage}%
}%
\subfigure[MAE on PEMS08]{
\begin{minipage}[t]{0.45\linewidth}
\centering
\includegraphics[width=1\linewidth]{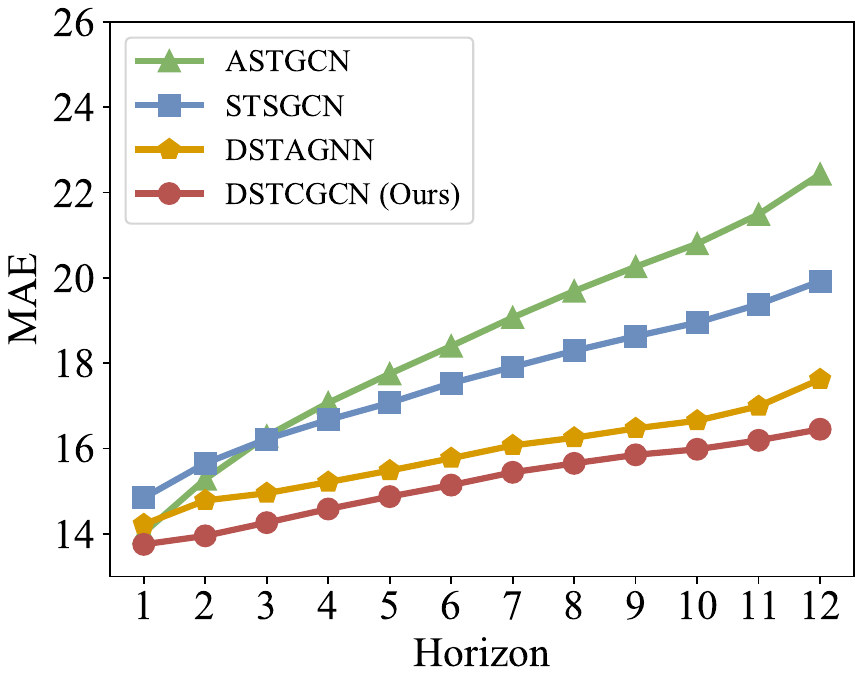}
\end{minipage}%
}%
\quad 
\subfigure[RMSE on PEMS04]{
\begin{minipage}[t]{0.45\linewidth}
\centering
\includegraphics[width=1\linewidth]{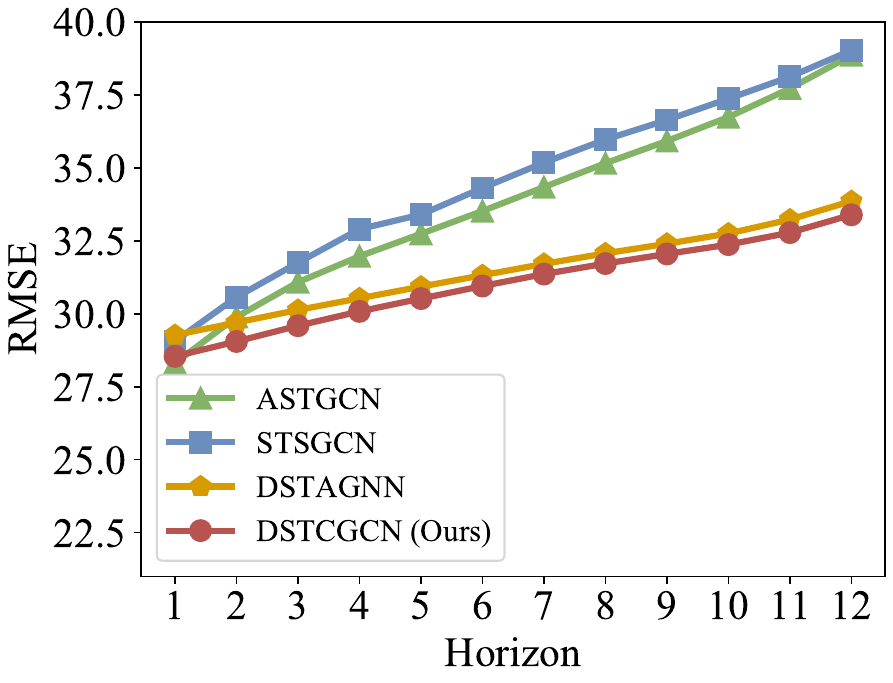}
\end{minipage}
}%
\subfigure[RMSE on PEMS08]{
\begin{minipage}[t]{0.45\linewidth}
\centering
\includegraphics[width=1\linewidth]{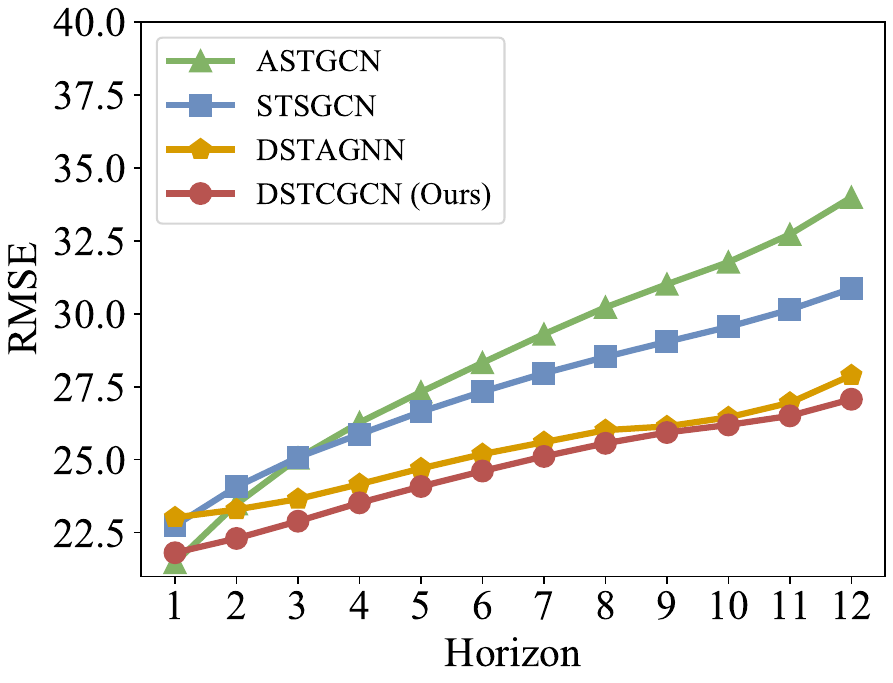}
\end{minipage}%
}%
\quad 
\subfigure[MAPE on PEMS04]{
\begin{minipage}[t]{0.45\linewidth}
\centering
\includegraphics[width=1\linewidth]{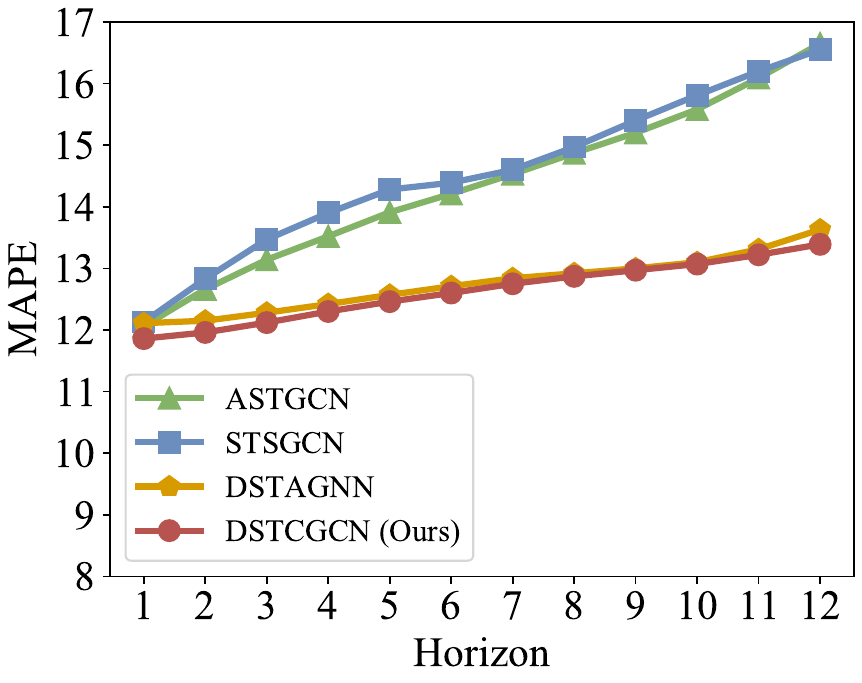}
\end{minipage}%
}%
\subfigure[MAPE on PEMS08]{
\begin{minipage}[t]{0.45\linewidth}
\centering
\includegraphics[width=1\linewidth]{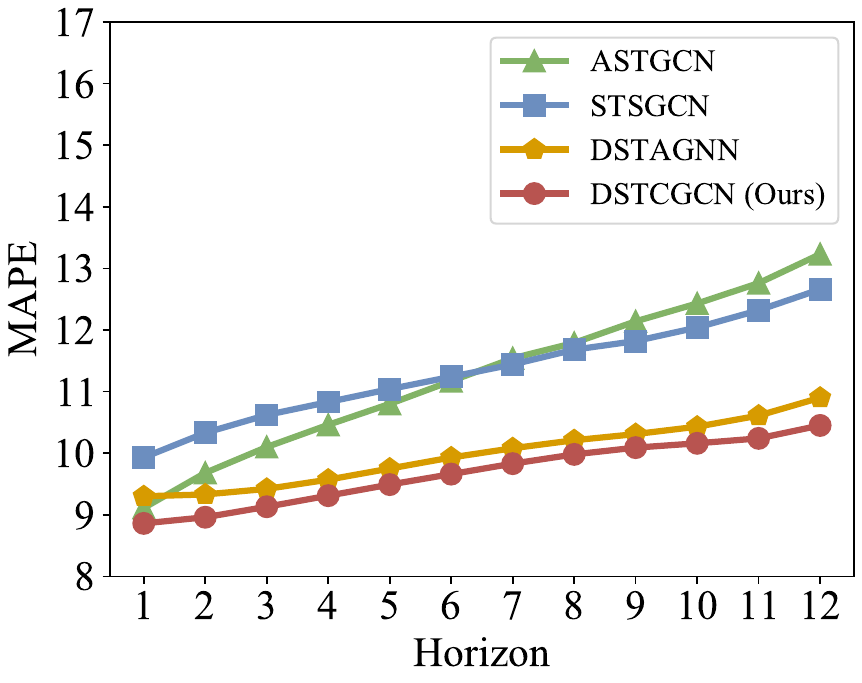}
\end{minipage}
}%
\caption{Forecasting performance comparison at each horizon on PEMS04 and PEMS08.}
\label{pre_pems_04_08}
\end{figure}

To further understand the effectiveness of critical hyperparameters in DSTCGCN, we conduct the hyperparameter evaluation. We utilize NNI to optimize our model. Considering the fairness, we fix other hyperparameters when investigating a certain hyperparameter.

We highlight the following two evaluations closely related to dynamic cross graphs:
\begin{itemize}
    \item Evaluation of the dimension of the node and time embeddings $d_\text{e}$. The node and time embeddings represent the characteristics of each node and each time step, respectively. $d_\text{e}$ significantly affects their representative abilities and influences the construction of dynamic spatial, temporal, and cross graphs. We increase $d_\text{e}$ from 5 to 17 with a step size of 1.
    \item Evaluation of the number of the selected relevant time steps $\tau$. $\tau$ determines the size of the spatial-temporal cross graphs, which controls the number of spatial and temporal neighbors when modeling cross dependencies. We increase $\tau$ from 1 to 4 with a step size of 1.
\end{itemize}

Fig. \ref{dim_04} shows the evaluation results of $d_\text{e}$ on PEMS04. It can be seen that increasing $d_\text{e}$ can improve the representative capacity of the model, leading to a drop in MAE. However, MAE rises quickly when it is larger than 10. The possible reason is that increasing dimensions brings more parameters to learn, making the model suffer from the over-fitting problem.

Fig. \ref{tau_04} shows the evaluation results of $\tau$ on PEMS04. It can be seen that both a small and large $\tau$ lead to weaker prediction performance and the optimal setting is 3. It may be because a small $\tau$ limits the number of spatial and temporal neighbors, which cannot fully discover the cross dependencies. Although a large $\tau$ has a large size of spatial-temporal neighbors, it may introduce noises into the model. Practically, $\tau=3$ on PEMS04 is suitable for DSTCGCN, as it balances the number of considered spatial-temporal neighbors and noises involved.

\subsection{Visualization}
Fig. \ref{pre_pems_04_08} shows the forecasting performance for all nodes at each horizon on PEMS04 and PEMS08, which demonstrates that DSTCGCN has a stable forecasting advantage in short-term and long-term forecasting.

\begin{figure}[]
\centering
\subfigure[Node \#129 in PEMS04]{
\begin{minipage}[t]{0.5\linewidth}
\centering
\includegraphics[width=\linewidth]{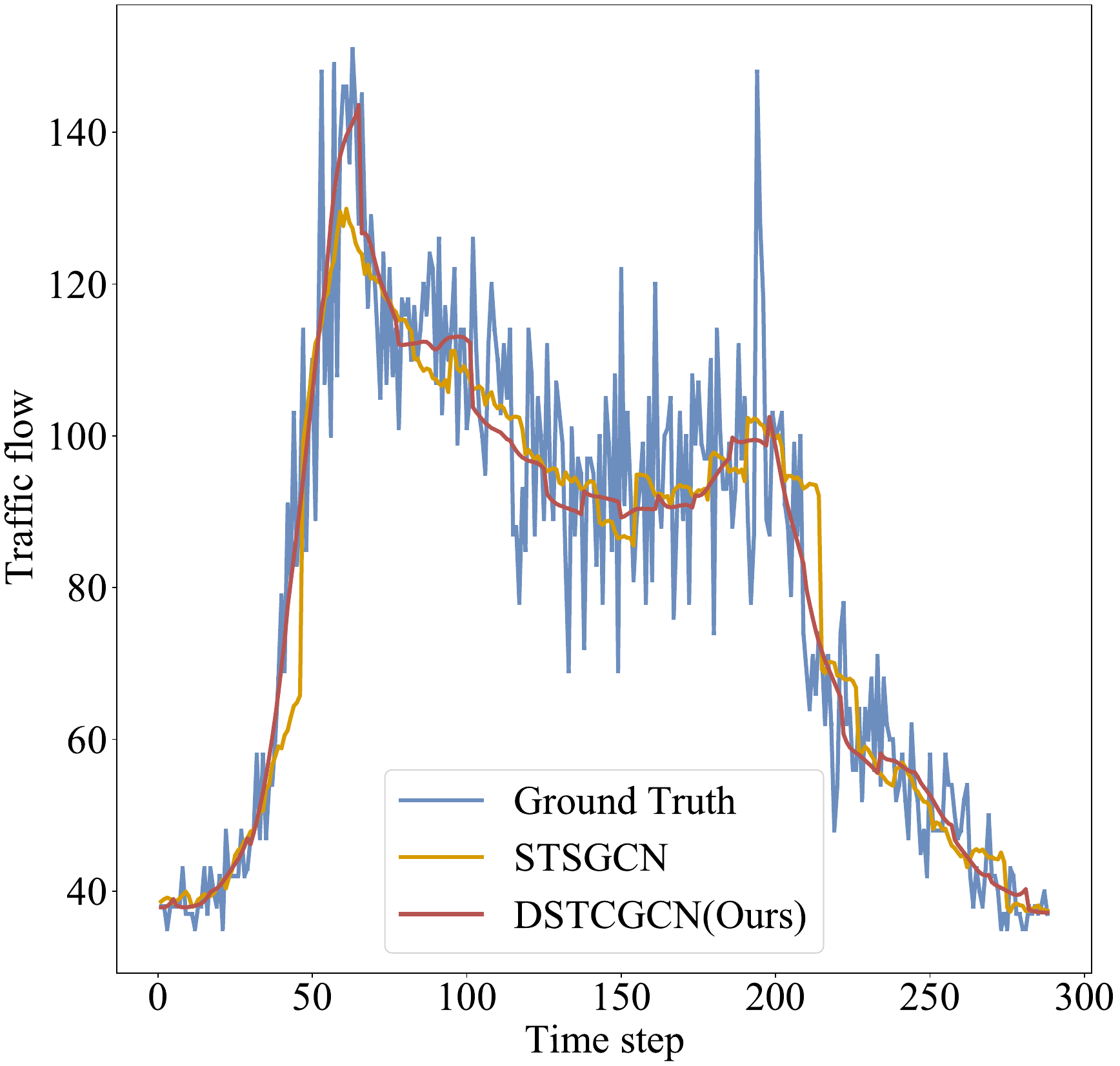}
\end{minipage}%
}%
\subfigure[Node \#196 in PEMS04]{
\begin{minipage}[t]{0.5\linewidth}
\centering
\includegraphics[width=\linewidth]{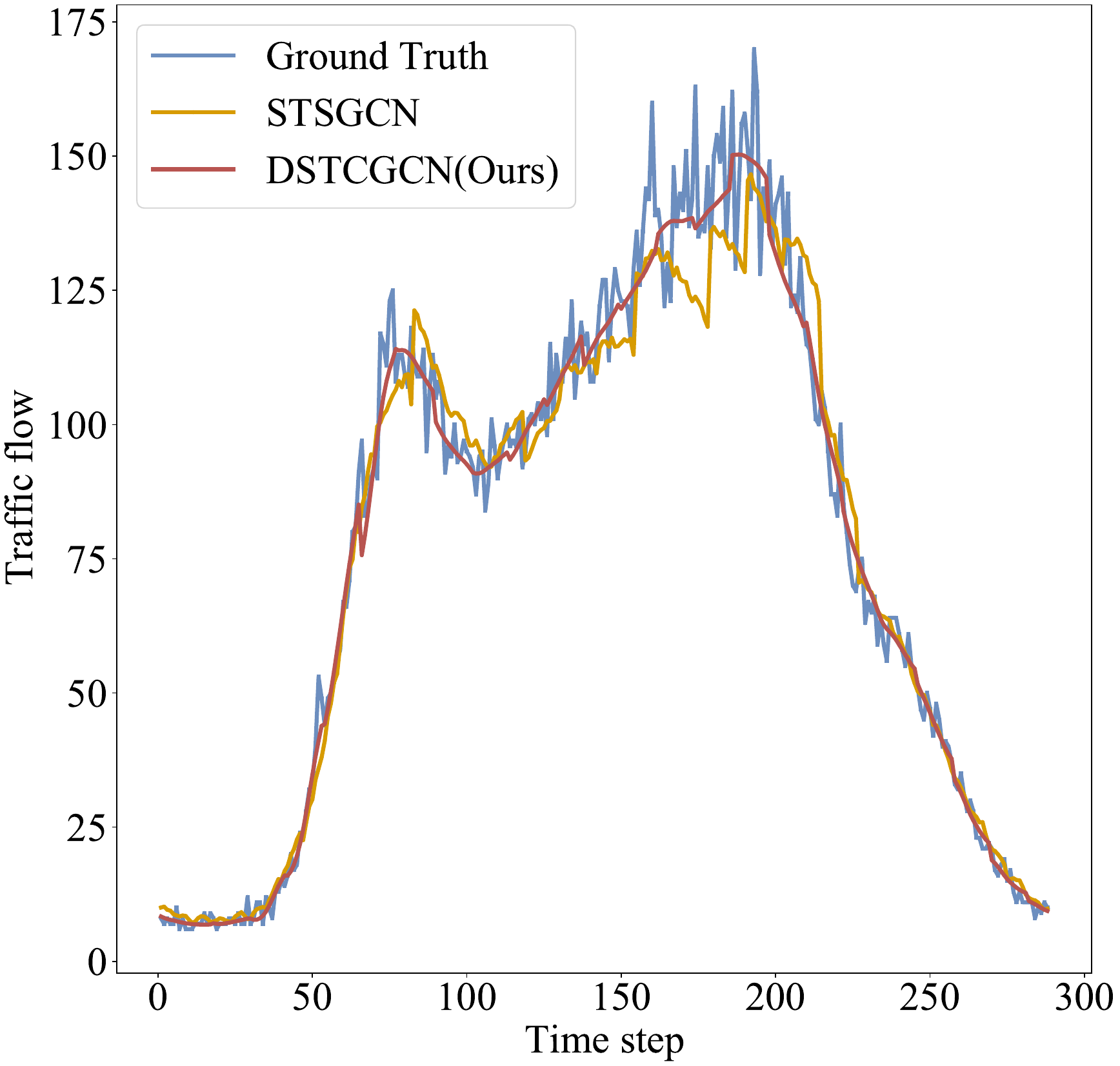}
\end{minipage}%
}%
\quad
\subfigure[Node \#69 in PEMS04]{
\begin{minipage}[t]{0.5\linewidth}
\centering
\includegraphics[width=\linewidth]{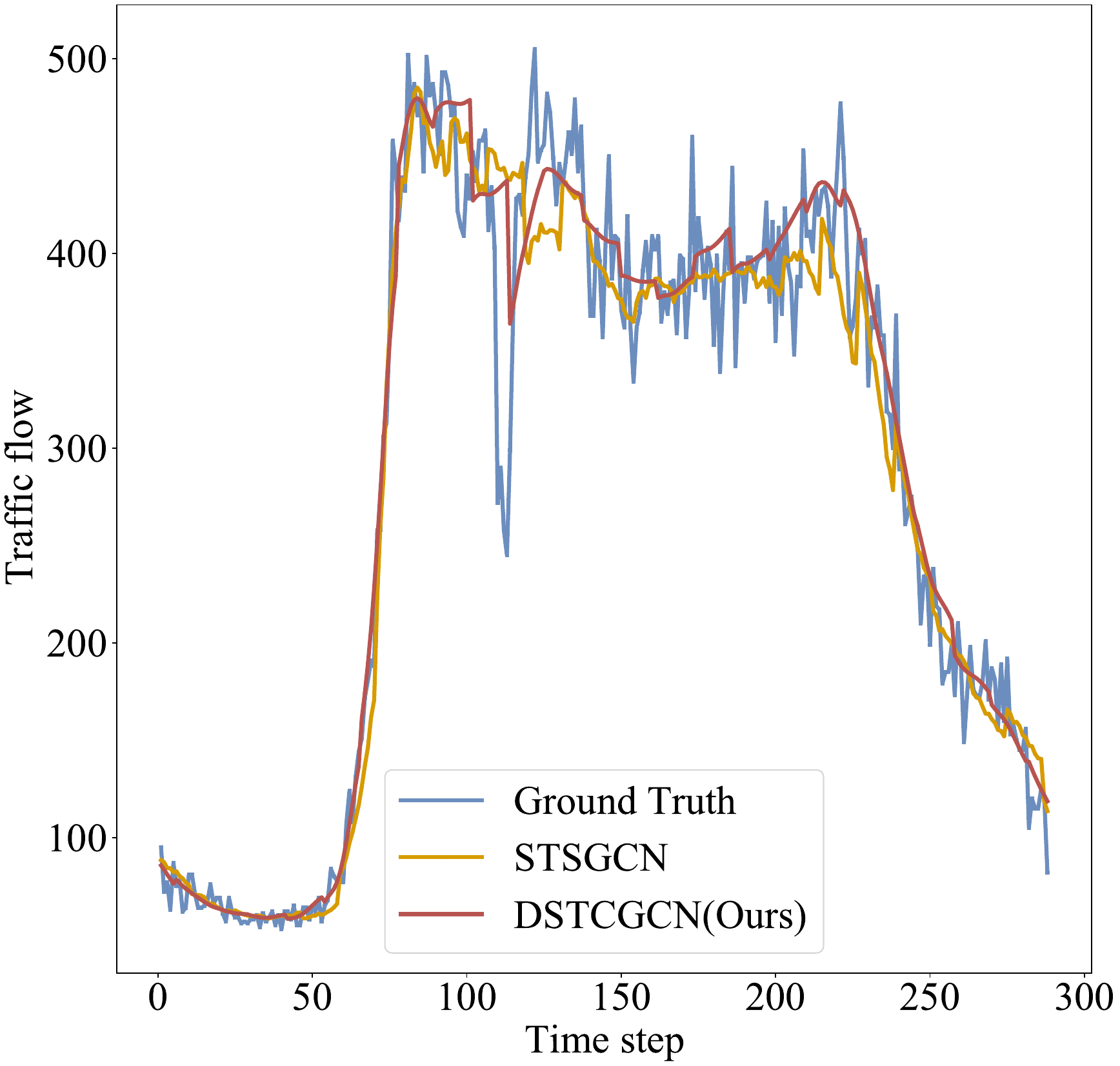}
\end{minipage}%
}%
\subfigure[Node \#89 in PEMS04]{
\begin{minipage}[t]{0.5\linewidth}
\centering
\includegraphics[width=\linewidth]{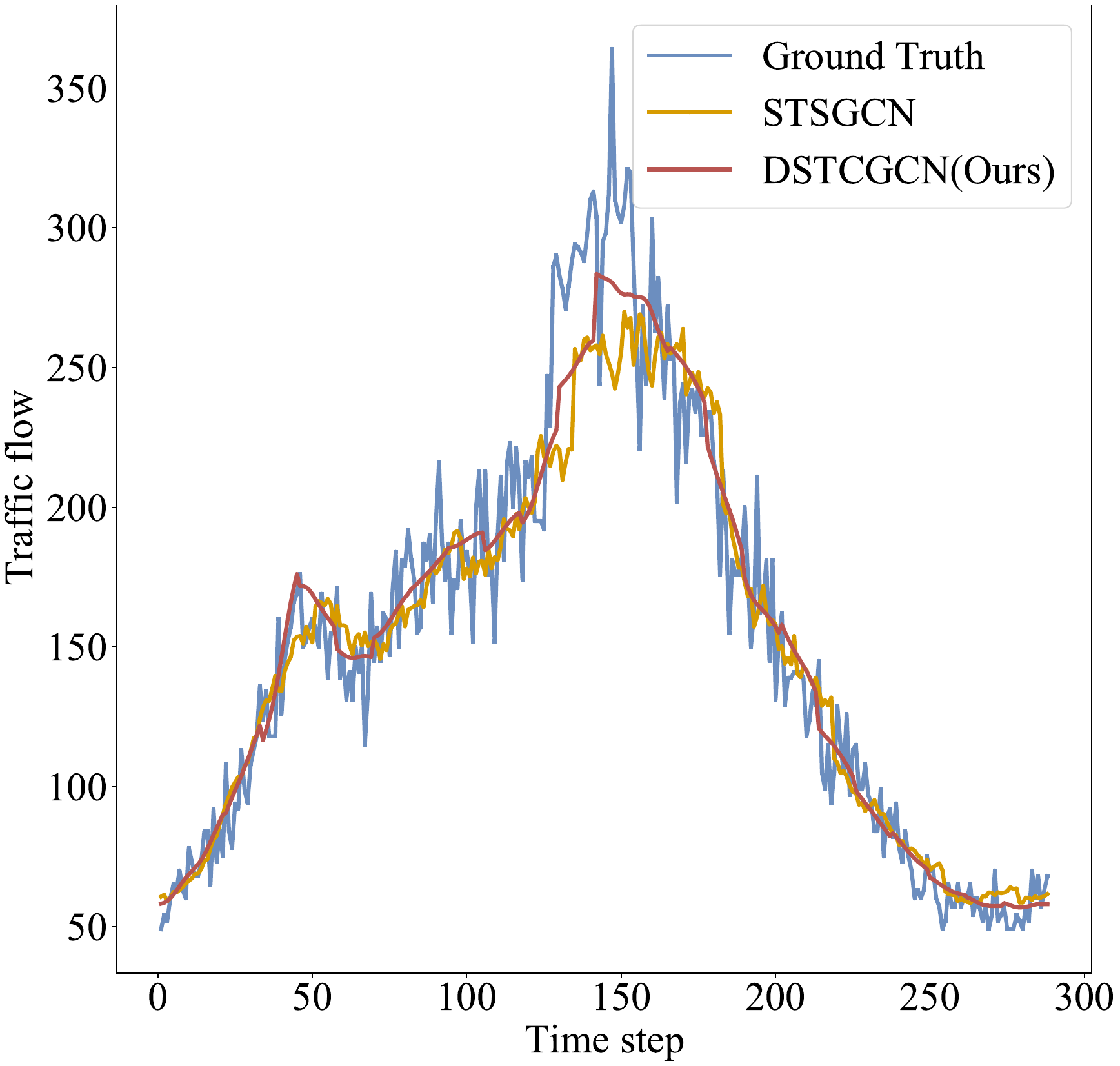}
\end{minipage}%
}%
\caption{Comparison of forecasting curves of Node \#129, \#196, \#69, and \# 89 between STSGCN and DSTCGCN on PEMS04.}
\label{case}
\end{figure}

To see the forecasting performance for specific nodes in the real-world cases, we randomly select one-day traffic flow data of 4 nodes in PMES04 and plot their ground truth and forecasting values of STSGCN and DSTCGCN. As shown in Fig. \ref{case}, DSTCGCN has a more quick and more accurate response to dynamic changes compared to STSGCN at the peak of the traffic flow curve. One of the possible reasons is that STSGCN utilizes a static distance matrix and static temporal connection graphs (identity matrix), which fails to model dynamic spatial and temporal dependencies jointly.

\begin{figure*}
    \centering
    \includegraphics[width=0.85\textwidth]{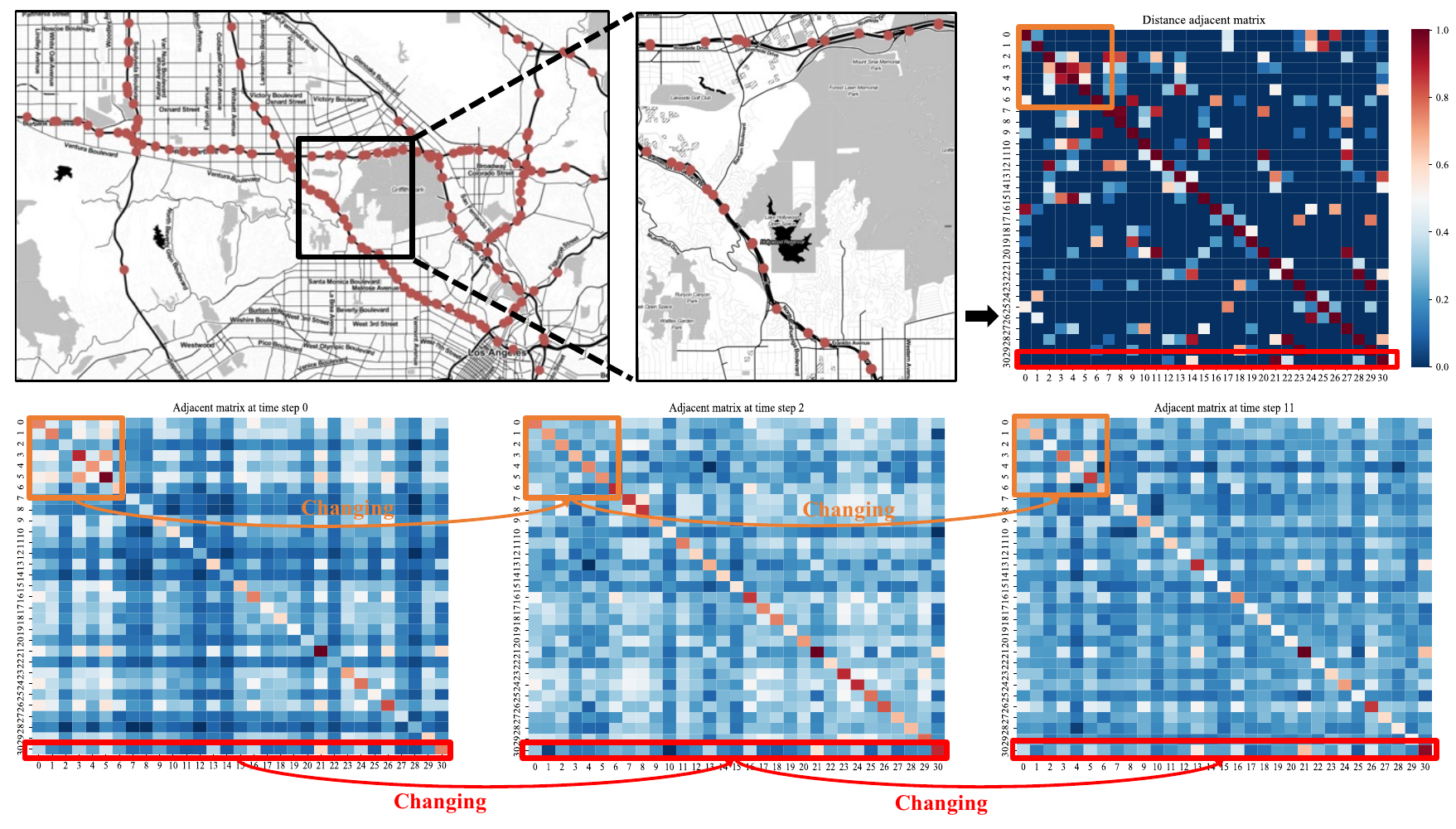}
    \caption{Illustration of the real-world distance adjacent matrix and learned dynamic spatial adjacent matrix for a case area on METR-LA.}
    \label{fig:dis}
\end{figure*}

To further illustrate the dynamic spatial and temporal dependencies modeling, we conduct case studies on METR-LA. We select an area with latitude from 34.10 to 34.16 and longitude from -118.30 to -118.36. The distribution of traffic speed sensors in that area and the corresponding distance adjacent matrix are illustrated in the above part of Fig. \ref{fig:dis}. We randomly select a period (12 continuous time steps) from the test dataset and display the learned dynamic spatial matrices at time steps 0, 2, and 11, as shown in the below part of Fig. \ref{fig:dis}. Comparing the dependencies between a node with others (see red windows) and within a sub-group of nodes (see orange windows) from the distance matrix and learned matrices, we find that our learned dynamic spatial matrices can not only capture the highly correlated geometric spatial dependencies but also can capture some dynamic hidden spatial dependencies.

\begin{figure*}
    \centering
    \includegraphics[width=0.85\textwidth]{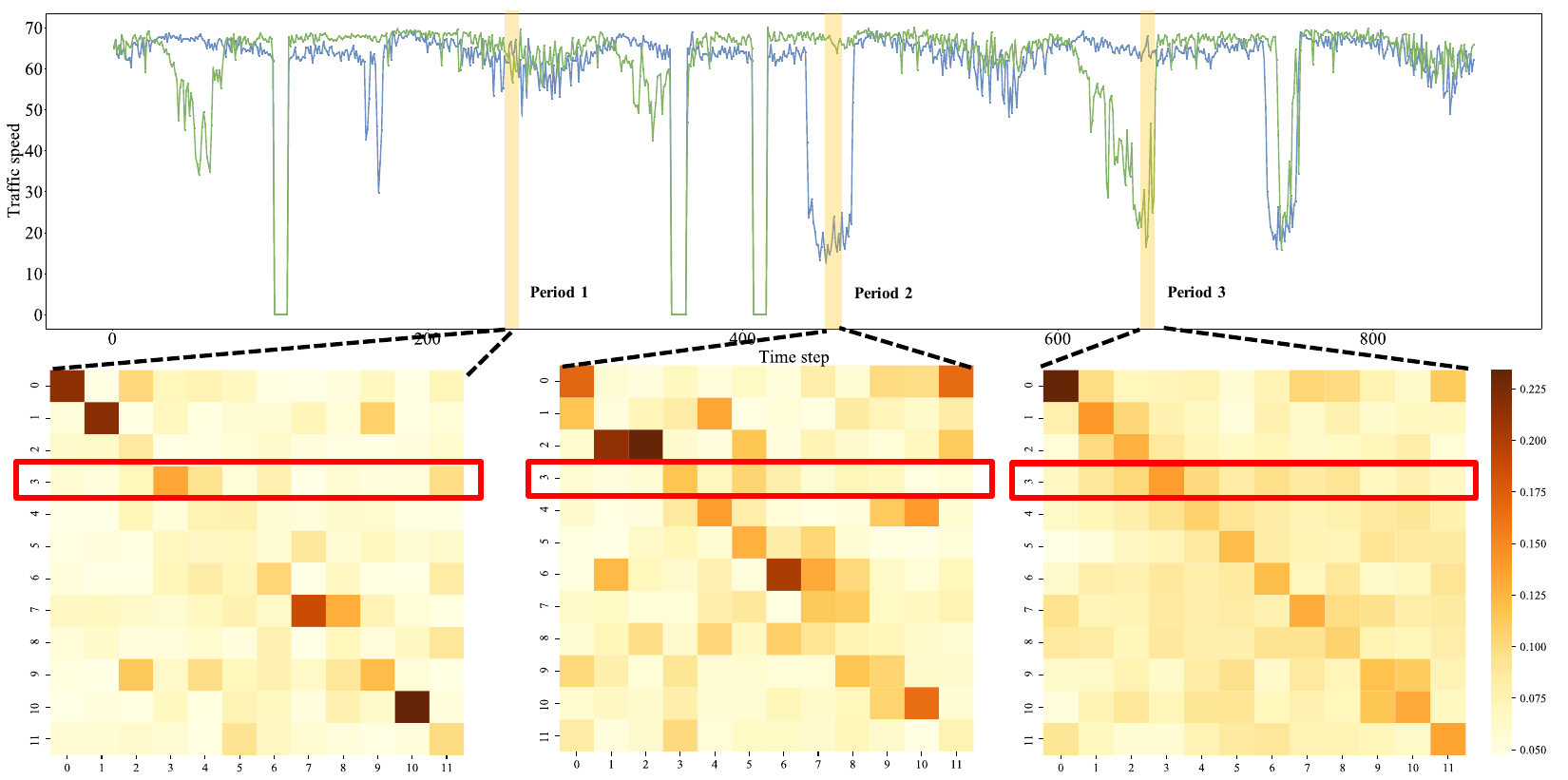}
    \caption{Illustration of the calculated FFT-attentive values for three case periods on METR-LA.}
    \label{fig:tim}
\end{figure*}

In addition, we randomly choose three periods from the test dataset and illustrate the FFT-attentive values that represent the temporal relevant weights between time steps in Fig. \ref{fig:tim}. For time step 3, the top-3 relevant time steps are 3, 11, and 4 for the period 1; 3, 5, and 6 for the period 2; and 3, 4, and 2 for the period 3. These phenomena support our motivation that the relevances between time steps are not static for all time, dynamic temporal dependencies should also be emphasized when modeling dynamics for traffic forecasting.

\subsection{Model Parameters and Training Time}
We compare the model parameters and training time of DSTCGCN with ASTGCN, RGSL, STSGCN, STFGCN, and TraverseNet on PEMS07 in Table \ref{comp_cost}. 
RGSL and DSTCGCN have impressive performance with relatively small amounts of model parameters and fast training time, as both of them adopt decomposition methods for graph convolution networks. In particular, DSTCGCN has an advantage in terms of accuracy compared with RGSL, as DSTCGCN can model temporal dependencies from a dynamic perspective, which matches the changing nature of the traffic system. The results show that DSTCGCN can achieve a good trade-off between computational cost and forecasting accuracy.

In addition, DSTCGCN has significant superiority compared with methods modeling cross dependencies, i.e., STSGCN, STFGCN, and TraveseNet. For example, due to the decomposition design, DSTCGCN has much fewer parameters than STSGCN and TraverseNet. The training time of DSTCGCN is almost half of that of TraverseNet, as DSTCGCN selects the top relevant time steps to model dynamic temporal dependencies rather than using all time steps as TraverseNet does. Such designs help DSTCGCN balance the computational cost and forecasting performance and reduce the risk of over-fitting when modeling cross dependencies.


\begin{table}[]
\caption{Computational cost of models on PEMS07.}
\label{comp_cost}
\centering
\begin{tabular}{c|ccc}
\hline
Methods & \# parameters & Training Time/epoch (min) & MAE \\ 
\hline
ASTGCN & 3,230,383 & 6.36 & 24.01 \\ 
RGSL & \textbf{879,482} & \textbf{2.91} & \underline{20.73}
\\ 
STSGCN & 8,340,861 & 5.52 & 25.51
\\ 
STFGCN & 968,204 & 5.12 & 23.46
\\ 
TraverseNet & 2,999,142 & 6.01 & 21.99
\\ \hline
DSTCGCN & \underline{900,858} & \underline{3.83} & \textbf{20.37}
\\ \hline
\end{tabular}
\end{table}

\section{Conclusions and Future Work}
In this paper, we propose DSTCGCN to learn dynamic spatial and temporal dependencies jointly via graphs for traffic forecasting. Specifically, we introduce 1) an FFT-based attentive selector to choose relevant time steps for each time step based on time-varying traffic data, and 2) a dynamic cross graph construction module to fuse time-varying spatial graphs and temporal connection graphs in a directed and sparse way. The results of our extensive experiments on six real-world traffic datasets show that DSTCGCN outperforms the state-of-the-art traffic forecasting baselines.

In the future, we focus on learning spatial-temporal cross dependencies with more casual properties. From Fig. \ref{fig:dis}, we find that the learned dynamic spatial matrices are dense, calling for a proper sparse strategy to regularize them and make the model more lightweight. Since causal dependencies have inherent sparsity, it is possible to introduce casual properties, e.g., Granger causality, when modeling spatial or cross dependencies. Moreover, some methods from explainable AI, e.g., ShapFlow \cite{Holzinger2022}, can be applied to analyze the causal effect of the learned dependencies, which can increase our understanding of traffic systems and provide reliable guidance for scheduling transportation resources.

\bibliographystyle{IEEEtran}
\bibliography{references.bib}
\vspace{-330pt}
\begin{IEEEbiography}[{\includegraphics[width=1in,height=1.25in,clip,keepaspectratio]{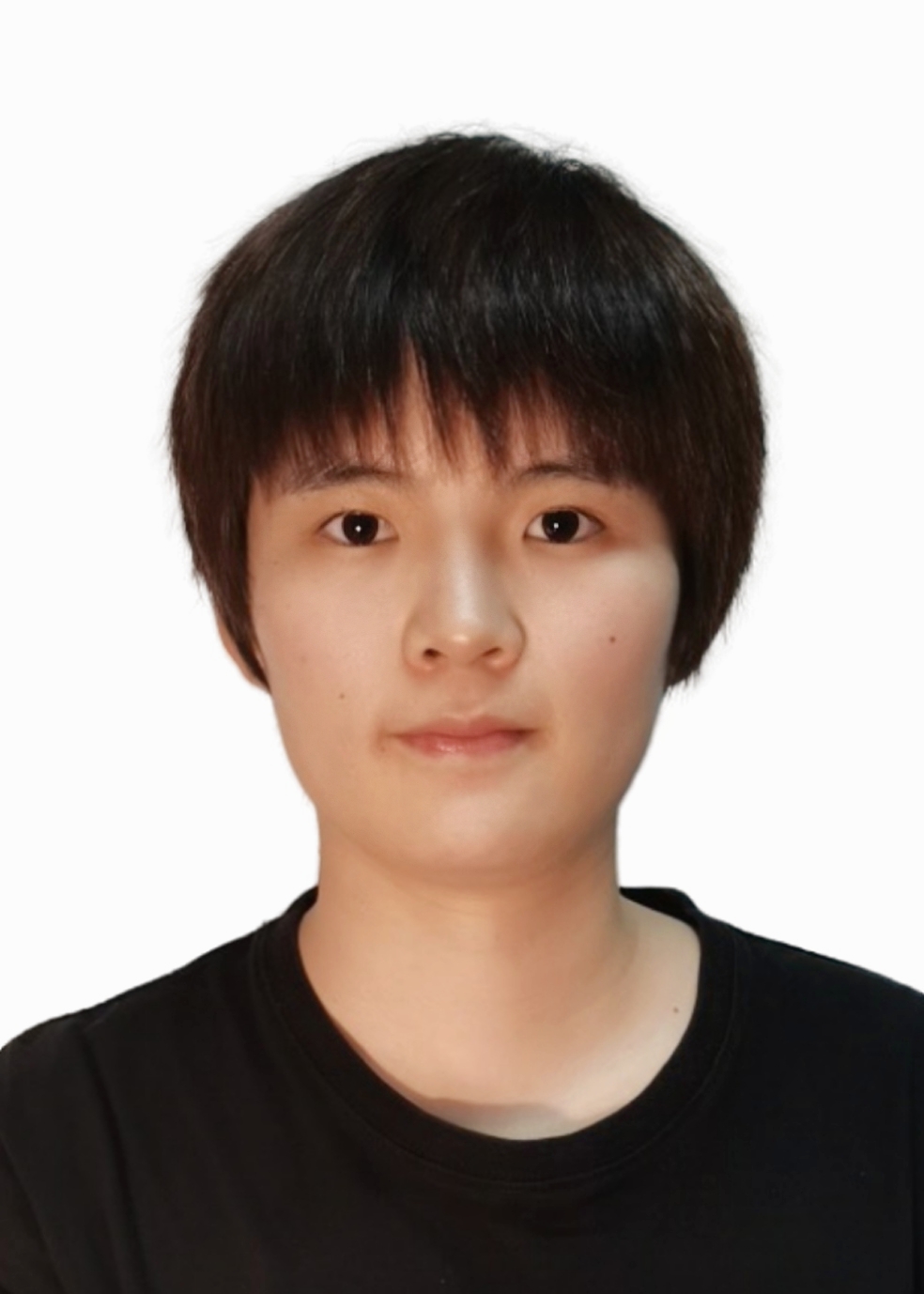}}]
{Binqing Wu} received her B.S. degrees in computer science from Southwest Jiangtong University, China, and University of Leeds, UK, in 2020. She is currently a Ph.D. candidate with the College of Computer Science and Technology, Zhejiang University, China. Her research interests include time series forecasting and data mining.
\end{IEEEbiography}

\vspace{-330pt}
\begin{IEEEbiography}[{\includegraphics[width=1in,height=1.25in,clip,keepaspectratio]{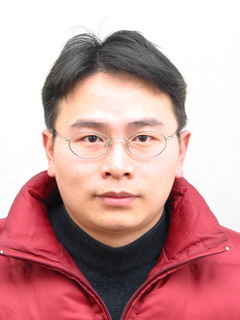}}]
{Ling Chen} received his B.S. and Ph.D. degrees in computer science from Zhejiang University, China, in 1999 and 2004, respectively. He is currently a professor with the College of Computer Science and Technology, Zhejiang University, China. His research interests include ubiquitous computing and data mining.
\end{IEEEbiography}


\end{document}